\documentclass{agujournal2019}
\usepackage{microtype}
\usepackage{lineno}
\usepackage{url}
\usepackage{soul}

\usepackage{natbib}
\usepackage{graphicx}
\usepackage{multirow}
\usepackage{amsmath,amssymb,amsfonts}
\usepackage{amsthm}
\usepackage{mathrsfs}
\usepackage[title]{appendix}
\usepackage{xcolor}
\usepackage{textcomp}
\usepackage{manyfoot}
\usepackage{booktabs}
\usepackage{algorithm}
\usepackage{algorithmicx}
\usepackage{algpseudocode}
\usepackage{listings}
\usepackage{svg}
\usepackage{subcaption}
\usepackage{array}
\usepackage[export]{adjustbox}
\newenvironment{plainlanguage}{%
  \par\medskip
  \noindent\textbf{Plain Language Summary}\par
  \smallskip
}{%
  \par\medskip
}

\draftfalse

\journalname{JGR: MLC}

\begin{document}
\sloppy   
\title{Incomplete Observations Boost Evolutionary Performance in Ocean Modeling}

\authors{
Yangyang Kong\textsuperscript{*,1,2},
Yutong Jiang\textsuperscript{*,1,2},
Yanhai Gan\affil{1,2},
Junyu Dong\affil{1,2},
Feng Gao\affil{1,2},
Xiaopei Lin\affil{1}
}

\begingroup
\renewcommand\thefootnote{*}
\footnotetext{These authors contributed equally to this work.}
\endgroup

\affiliation{1}{State Key Laboratory of Physical Oceanography, Ocean University of China, Qingdao, China}

\affiliation{2}{School of Computer Science and Technology, Ocean University of China, Qingdao, China}

\correspondingauthor{Yanhai Gan}{ganyanhai@ouc.edu.cn}
\correspondingauthor{Junyu Dong}{dongjunyu@ouc.edu.cn}



\begin{keypoints}
  \item A generative model integrating physical states and observations is proposed, following the paradigm of conventional numerical models.
  \item An optimization framework is formulated, enabling the generative model to directly learn ocean dynamics from sparse observations.
  \item Leveraging sparse observations, the model realizes self-evolution and yields improved reconstruction and prediction performance.
\end{keypoints}

%
%

%
%


\begin{abstract}

Data-driven methods have revolutionized ocean modeling, yet current approaches rely heavily on complete reanalysis datasets, imposing computational constraints and limiting model performance to that of the training data. Here, we present a generative state-space model and an optimization framework that enable learning directly from sparse and noisy observations. The model is essentially a hidden Markov model with a continuous state space, where oceanic physical quantities are treated as hidden states and measurements as observations, enabling a unified representation of ocean fields and observational data. Both the initial-state and state-transition modules are implemented as neural networks to capture the complexity and temporal evolution of ocean states, while the emission module is formulated as a masked Gaussian distribution. To train the model from sparse observations, we derive an optimization framework based on the expectation–maximization (EM) algorithm. The framework alternately reconstructs high-fidelity ocean fields via Langevin dynamics and optimizes deep neural networks to capture temporal evolution. Theoretical analysis shows that the framework maximizes the likelihood of observations under the generative model. For efficiency, we assume that ocean-state evolution follows a stationary, ergodic, and Markovian stochastic process and adopt only length-two state sequences during optimization. Experiments on CMIP6 simulation data and FY-3D satellite data demonstrate high-fidelity reconstruction and accurate prediction, showing that sparse observations can directly improve the model’s representation of ocean-state dynamics. This work offers a scalable pathway for next-generation Earth system models to learn directly from sparse, incomplete real-world observations.
\end{abstract}

\begin{plainlanguage}
Today's most successful artificial intelligence models for simulating the ocean and atmosphere are trained on large, carefully reconstructed global datasets that combine sparse observations with traditional computer models. While these datasets are useful, they are expensive to produce, limit the finest scales that can be resolved, and cap how accurate artificial intelligence predictions can become. In this study, we developed a new machine learning framework that can learn directly from the incomplete, sparse, and noisy measurements collected by real-world instruments such as satellites. Our approach treats the complete ocean state as a quantity that evolves over time and uses an iterative training strategy: first, it fills in missing observations to build realistic ocean fields, then it trains neural networks to predict how those fields change. Tests using climate model simulations and sea surface temperature measurements from a Chinese weather satellite show that this method improves prediction accuracy and can rebuild complete ocean maps even when much of the input data are missing. These results suggest a practical path toward artificial intelligence Earth system models that learn from real observations rather than depending solely on reconstructed datasets.
\end{plainlanguage}

\begingroup
\renewcommand\thefootnote{*}
\footnotetext{These authors contributed equally to this work.}
\endgroup

\section{Introduction}

In recent years, artificial intelligence (AI), spearheaded by deep learning, has driven a substantial shift in scientific discovery, with growing implications for the Earth sciences under the banner of ``AI for Science'' \citep{Reichstein2019Nature}.
The emergence of next-generation data-driven foundation models, exemplified by Pangu-Weather \citep{Bi2023Nature}, GraphCast \citep{Lam2023Science}, FengWu \citep{Chen2023FengWu}, and FourCastNet \citep{Pathak2022FourCastNet}, is reshaping Earth system modeling.
These models achieve accuracies comparable to traditional numerical weather prediction (NWP) at substantially lower inference cost, representing a notable advance \citep{Nguyen2023ClimaX}.
These developments have made the goal of constructing a ``Digital Twin'' capable of simulating and predicting complex Earth systems at high resolution and fidelity more attainable than before.

However, the success of these state-of-the-art AI models rests almost invariably on large-scale, high-quality reanalysis datasets, such as ERA5 for the atmosphere \citep{Hersbach2020QJRMS} and global ocean state estimates including GLORYS and ECCO \citep{Forget2015ECCO}.
Reanalysis fields are produced by fusing sparse, multi-modal observations with simulations from traditional physical numerical models via data assimilation (DA).
Although such products are currently the principal training data, dependence on this complete-field paradigm imposes two constraints on further progress.
First, the computational cost of ocean data assimilation limits the spatiotemporal resolution of reanalysis products, thereby leaving finer, dynamically important sub-mesoscale processes under-resolved relative to the capacity of AI models \citep{Brajard2020JCS}.
Second, reanalysis accuracy is bounded by observational coverage and numerical model fidelity, creating an effective performance ceiling associated with intrinsic uncertainties.
When AI models treat reanalysis data as ground truth, predictive skill is largely limited by reanalysis fidelity itself, which hinders their development into independent physical simulators that could, in principle, complement traditional numerical methods.
Consequently, a central scientific question is whether dependence on reanalysis can be reduced and whether a training paradigm can be established that allows AI models to learn directly from incomplete, sparse, and noisy ocean observations---such as cloud-contaminated satellite swath measurements with large spatial gaps \citep{Zhang2025JGRMLC}.
Meeting this challenge may move AI models beyond emulating assimilated products toward learning observation-consistent dynamics \citep{Runge2019NatComms}.
Yet existing generative paradigms also face difficulties.
Whether based on generative adversarial networks (GANs) \citep{Ravuri2021Nature}, diffusion models \citep{Mardani2023CorrDiff}, or recent sparse reconstruction methods, these approaches encounter fundamental limitations when applied directly to such scientific tasks.
Designed primarily for natural image synthesis, standard generative models are typically not constructed with explicit physical constraints.
Although they can produce visually plausible static fields, they often struggle to maintain physical consistency over complex temporal evolution---a gap increasingly noted in Earth system machine learning \citep{Sonnewald2019,Feng2025JGRMLC}.
In particular, they lack explicit mechanisms to preserve approximate geostrophic balance, mass conservation, or large-scale circulation features characteristic of ocean dynamics.

To address this gap, we propose a generative state-space modeling framework designed for incomplete observational data.
By incorporating a state-space representation of temporal dynamics, the framework maintains spatiotemporal coherence within the generative process.
The complete ocean field is treated as a latent state, heterogeneous measurements are linked through a masked Gaussian observation model, and deep generative networks parameterize the initial-state distribution and stochastic state transitions.
Parameters are estimated using an expectation--maximization (EM) strategy \citep{Dempster1977EM}.
In the expectation (E) step, we use Monte Carlo sampling based on Langevin dynamics; conditioned on sparse observations, this step reconstructs high-fidelity, spatiotemporally continuous oceanographic fields.
In the maximization (M) step, the reconstructed fields supervise updates to the initial-state generator (a modified StyleGAN2-ADA backbone) and the state-transition model (a stochastic U-Net) \citep{Ross2023JAMES}.
Iterating between the E- and M-steps allows the model to refine physical patterns from noisy and missing data and to reconstruct and predict spatiotemporal evolution of the physical system.

\section{Methodology}
\label{sec:methodology}

\subsection{Probabilistic formulation as a Generative State-Space Model}
\label{subsec:ssm_formulation}
To address the challenge of learning Earth system dynamics from incomplete observations~\citep{Reichstein2019Nature, Ghil1991, Brajard2020JCS}, we formulate the problem within a probabilistic generative modeling framework. Specifically, we structure the system's dynamics as a Generative State-Space Model (SSM)~\citep{Deep_SSM}, where the complete physical fields, $\mathbf{s} = \{\mathbf{s}_0, \dots, \mathbf{s}_T\}$, are treated as state variables and the sparse satellite data, $\mathbf{o} = \{\mathbf{o}_0, \dots, \mathbf{o}_T\}$, as observed variables. By assuming a first-order Markov process, the joint probability distribution over the entire system, $p_\theta(\mathbf{s}, \mathbf{o})$, can be factorized into three core components: an Initial State Model, a State Transition Model, and an Observation Model. Within this framework, our fundamental optimization objective is maximum likelihood estimation (MLE), which seeks parameters $\theta^*$ that maximize the marginal likelihood of the observed data $p_\theta(\mathbf{o})$. Due to the latent nature of $\mathbf{s}$, this requires an intractable high-dimensional integral over all state configurations:
\begin{equation}
    \theta^* = \arg\max_{\theta} \log p_\theta(\mathbf{o}) = \arg\max_{\theta} \log \int p_\theta(\mathbf{s}, \mathbf{o}) \, \mathrm{d}\mathbf{s}.
    \label{eq:mle_objective}
\end{equation}

\subsection{Learning via Monte Carlo Expectation-Maximization}
\label{subsec:mcem_learning}
Direct optimization of the MLE objective in Eq.~\eqref{eq:mle_objective} is intractable.
To address this, we adopt the Expectation--Maximization (EM) framework~\citep{Dempster1977EM} to iteratively approach Eq.~\eqref{eq:mle_objective}.
Since the expectations in the M-step are computationally prohibitive, we implement a Monte Carlo EM (MCEM) approach~\citep{MCEM_1990, Peyron2021}.
In the E-step, we draw one state trajectory (i.e., a realization of $\mathbf{s}$) using Langevin dynamics to approximate the posterior $p_{\theta^{(t)}}(\mathbf{s} \mid \mathbf{o})$ given observations.
For a dataset comprising $N$ samples (i.e., distinct realizations of $\mathbf{o}$), the M-step updates the parameters by solving
\begin{equation}
    \theta^{(t+1)} = \arg\max_{\theta} \sum_{i=1}^{N} \log p_\theta\bigl(\mathbf{s}^{(i)}, \mathbf{o}^{(i)}\bigr).
    \label{eq:mcem_objective}
\end{equation}
Here, $\mathbf{s}^{(i)}$ is imputed in the E-step at iteration $t$ and is strictly paired with $\mathbf{o}^{(i)}$.
The superscript ``$(i)$'' indexes different instances in the dataset.
Eq.~\eqref{eq:mcem_objective} is the M-step update rule in the same $\arg\max$ form as Eq.~\eqref{eq:mle_objective}; the optimization objective remains maximization of $\log p_\theta(\mathbf{o})$ in Eq.~\eqref{eq:mle_objective}.

Detailed mathematical derivations, including the decomposition of the evidence lower bound and the Monte Carlo rationale, are provided in Appendix~A.

This formulation effectively transforms the original unsupervised learning problem into a standard supervised learning task, where the M-step updates $\theta$ by solving Eq.~\eqref{eq:mcem_objective}, while the overall learning goal remains Eq.~\eqref{eq:mle_objective}.

\subsection{Deep generative architectures}
\label{subsec:architectures}

The probabilistic components of our state-space model are parameterized by advanced deep neural networks, designed to balance generation quality with computational efficiency and training stability.

\subsubsection{Initial state model}
\label{sssec:initial_state}
We parameterize the initial state distribution $p_{\theta_{init}}(\mathbf{s}_0)$ using a modified StyleGAN2-ADA architecture ~\citep{Karras2020ada}. While StyleGAN2-ADA provides robust training on limited scientific datasets via adaptive discriminator augmentation, its standard convolutional blocks are computationally expensive. To achieve a lightweight design, we replace standard convolutions with Depthwise Separable Convolutions (DSC) ~\citep{Chollet2017}. However, applying the standard StyleGAN modulation-demodulation mechanism directly to depthwise convolution (DWConv) kernels leads to a mathematical cancellation effect, where the style scaling factors act as both numerator and denominator during normalization, effectively erasing style information due to the channel independence of DWConv.

To resolve this, we introduce a structural decoupling strategy for style injection. Specifically, we restrict the style modulation operation to the DWConv layer, allowing it to exclusively adjust the relative amplitudes of individual feature channels. Subsequently, we utilize the Pointwise Convolution (PWConv) layer to linearly mix these style-modulated features, transforming amplitude differences into structural variations. Crucially, the demodulation (normalization) operation, originally performed at the DWConv layer, is migrated to the PWConv layer. This ``inject-at-DWConv, normalize-at-PWConv'' design avoids mathematical cancellation, ensuring effective style propagation across scales while significantly improving parameter efficiency.

\subsubsection{Stochastic state transition model}
\label{sssec:transition_model}
The transition probability $p_{\theta_{trans}}(\mathbf{s}_{t+1}|\mathbf{s}_t)$ is learned via a conditional adversarial framework inspired by Pix2Pix ~\citep{isola2017image}. The generator employs a Stochastic U-Net architecture  ~\citep{Ronneberger2015}, which takes the concatenation of the current state $\mathbf{s}_t$ and a latent noise tensor $\mathbf{z}$ as input. The U-Net's multi-scale skip connections efficiently capture the multi-resolution spatial features inherent in ocean-meteorological systems. Unlike the local PatchGAN discriminator used in standard Pix2Pix, we employ a global conditional convolutional classifier as the discriminator to enforce global physical consistency constraints ~\citep{mirza2014conditional}, a strategy increasingly emphasized in recent data-driven Earth system modeling to prevent unphysical predictions.

To enable probabilistic forecasting and enhance optimization flexibility, we incorporate an explicit noise injection mechanism. Gaussian noise sampled from $\mathcal{N}(0, \mathbf{I})$ is injected into each layer of the decoder after passing through learned affine transformations ~\citep{StyleGAN}. This design provides dual benefits: it transforms the deterministic mapping into a stochastic process to capture the intrinsic uncertainty of physical evolution, and it expands the solution space flexibility, providing critical manifold support for the gradient-based posterior sampling in the E-step.

\subsubsection{Observation model}
\label{sssec:observation_model}
    We establish an explicit probabilistic observation model $p(\mathbf{o}_t|\mathbf{s}_t)$ to mathematically describe the link between the hidden system state $\mathbf{s}_t$ and the incomplete observation $\mathbf{o}_t$. We assume that observations are measurements taken from a sparse spatial subset of the true physical field, corrupted by Gaussian noise. This is formalized as a conditional Gaussian distribution, consistent with variational data assimilation principles in oceanography ~\citep{Kalnay2003, Evensen2009, Bennett2002}:
\begin{equation}
    p(\mathbf{o}_t|\mathbf{s}_t) = \mathcal{N}(\mathbf{o}_t \mid \mathbf{M}_t \odot \mathbf{s}_t, \sigma^2\mathbf{I}).
\end{equation}
where $\mathbf{M}_t$ is a binary mask operator representing the spatial location of observations (e.g., determined by satellite orbits or cloud cover in real-world datasets, or generated artificially in simulated datasets), and $\sigma^2\mathbf{I}$ represents the measurement uncertainty.

\subsection{Iterative Optimization via Expectation-Maximization}
\label{subsec:em_optimization}

In practice, directly optimizing the log-likelihood in Eq.~\eqref{eq:mle_objective} is extremely challenging. On one hand, explicitly modeling the joint probability of state and observation variables is difficult. On the other hand, computing the marginal likelihood requires integrating over the latent state space, which is intractable even if the joint probability is known. Intuitively, constructing a model from incomplete observations faces a dilemma of circular dependency between ``state reconstruction'' and ``model learning'' ~\citep{Ghahramani1999}: the precise inference of the complete physical field $\mathbf{s}$ requires an accurate model as a prior, yet training an accurate model relies on a comprehensive representation of the complete field $\mathbf{s}$.

To address this circular dependency, we employ the Expectation-Maximization (EM) algorithm~\citep{Dempster1977EM}, transforming the problem into two alternating core steps. The fundamental objective remains to maximize the observation likelihood $p_\theta(\mathbf{o})$. In the E-step, we obtain samples from the posterior distribution $p_\theta(\mathbf{s}|\mathbf{o})$ via Langevin dynamics to achieve an efficient reconstruction of the system state, a strategy increasingly adopted in data-driven oceanography~\citep{Fablet2021, Bocquet2019}. In the subsequent M-step, these reconstructed state samples are used as high-quality training data to update the parameters of the joint probability model. The E-step and M-step alternate, driving the iterative evolution of the entire framework.

Our proposed framework is theoretically applicable to sequential problems of arbitrary length $T$, offering a generalized solution for learning spatiotemporal dynamics from incomplete data. The detailed algorithmic flow of a single EM iteration is summarized in Algorithm~\ref{alg:em_iteration}.

\begin{algorithm}[h!] 
\caption{One Iteration of the EM-based Learning Framework}\label{alg:em_iteration}
\begin{algorithmic}[1]
\Require \textbf{Models:} Initial state generator $G_{\theta_{\text{init}}}$, Transition model $F_{\theta_{\text{trans}}}$.
\Require \textbf{Data:} Sparse observations $\{\mathbf{o}_t\}_{t=0}^T$, Mask operators $\{\mathbf{M}_t\}_{t=0}^T$.
\Require \textbf{Hyperparameters:} Langevin steps $K$, Step size $\eta$, Prior weights $\Lambda=\{\lambda_{\mathbf{z}}\}$.
\Ensure Updated parameters $\theta^{\text{new}} = \{\theta_{\text{init}}, \theta_{\text{trans}}\}$.

\Statex
\Statex \textbf{E-Step: State Reconstruction via Latent Space Sampling}
\State Initialize latent variables $\mathcal{Z}=\{\mathbf{z}_0, \dots, \mathbf{z}_T\} \sim \mathcal{N}(0,\mathbf{I})$.

\For{$k=1$ to $K$}
    \State \textit{Forward simulation to generate physical states:}
    \State $\mathbf{s}_0 \leftarrow G_{\theta_{\text{init}}}(\mathbf{z}_0)$
    \For{$t=1$ to $T$}
        \State $\mathbf{s}_t \leftarrow F_{\theta_{\text{trans}}}(\mathbf{s}_{t-1}, \mathbf{z}_t)$
    \EndFor

    \State \textit{Compute energy (negative log-posterior):}
    \State $\mathcal{E}(\mathcal{Z}) \leftarrow \sum_{t=0}^{T} \frac{\| (\mathbf{o}_t - \mathbf{s}_t) \odot \mathbf{M}_t \|_2^2}{2\sigma_t^2} + \sum_{\mathbf{z}\in\mathcal{Z}} \lambda_{\mathbf{z}} \|\mathbf{z}\|_2^2$

    \State \textit{Update latents via Langevin Dynamics step:}
    \State $\mathcal{Z} \leftarrow \mathcal{Z} - \frac{\eta}{2}\nabla_{\mathcal{Z}} \mathcal{E}(\mathcal{Z}) + \sqrt{\eta}\cdot\mathcal{N}(0,\mathbf{I})$
\EndFor
\State \textbf{Output of E-Step:} Reconstructed full states $\hat{\mathcal{S}}$ derived from the final $\mathcal{Z}$.

\Statex
\Statex \textbf{M-Step: Model Parameter Updates}
\State \textit{Optimize Initial-State Model:}
\State $\mathcal{L}_{\text{init}} \leftarrow \sum_{t=0}^{T} \mathcal{L}_{G}(\theta_{\text{init}}; \hat{\mathbf{s}}_t)$
\State $\theta_{\text{init}} \leftarrow \text{AdamUpdate}(\theta_{\text{init}}, \nabla_{\theta_{\text{init}}}\mathcal{L}_{\text{init}})$

\Statex
\State \textit{Optimize Transition Model:}
\State $\mathcal{L}_{\text{trans}} \leftarrow \sum_{t=1}^{T} \mathcal{L}_{F}(\theta_{\text{trans}}; \hat{\mathbf{s}}_{t-1}, \hat{\mathbf{s}}_t)$
\State $\theta_{\text{trans}} \leftarrow \text{AdamUpdate}(\theta_{\text{trans}}, \nabla_{\theta_{\text{trans}}}\mathcal{L}_{\text{trans}})$

\Statex
\State \textbf{return} Updated parameters $\theta^{\text{new}}$.
\end{algorithmic}
\end{algorithm}

\subsection{E-step: Efficient state reconstruction via latent space sampling}
\label{subsec:estep}

This step addresses the core challenge of state reconstruction within our framework, specifically aiming to ``achieve efficient reconstruction of spatiotemporal system states conditional on observational data.'' Within the EM framework, the mathematical objective of the E-step is to sample from the posterior distribution $p_\theta(\mathbf{s}|\mathbf{o})$, given the current model parameters $\theta$ and the incomplete observations $\mathbf{o}$.

Directly performing Langevin dynamics sampling on the posterior in the pixel space ($\mathbf{s}$-space) faces significant challenges: the dimensionality of $\mathbf{s}$ is extremely high, and its probability distribution can be exceedingly complex and rugged. To address this, we adopt a more efficient and stable latent space sampling strategy~\citep{Latent_EBM}. The core idea is to shift from directly sampling the high-dimensional state $\mathbf{s}$ to sampling the lower-dimensional stochastic latent vector $\mathbf{z}$ that governs the generation of $\mathbf{s}$, which has shown promising results in inverting complex geophysical systems~\citep{Peyron2021}.

Specifically, we view the system state as being generated by our pre-trained state models from a sequence of latent noise vectors. Here, the latent noise vectors $\mathbf{z}$ encompass both the input noise and the layer-wise injected noise within the Initial State Model and the State Transition Model. Consequently, our objective transforms from sampling $\mathbf{s}$ to sampling $\mathbf{z}$:
\begin{equation}
    \mathbf{z} \sim p_\theta(\mathbf{z}|\mathbf{o}).
\end{equation}
Since $p_\theta(\mathbf{z}|\mathbf{o})$ is intractable to compute, we utilize Langevin dynamics for sampling~\citep{Welling2011}. To do this, we need to compute the score of the noise posterior distribution:
\begin{align}
    \nabla_{\mathbf{z}} \log p_\theta(\mathbf{z}|\mathbf{o}) &= \nabla_{\mathbf{z}} \log p_\theta(\mathbf{z}, \mathbf{o}) - \nabla_{\mathbf{z}} \log p_\theta(\mathbf{o}) \notag \\
    &= \nabla_{\mathbf{z}} \log p(\mathbf{z}) + \nabla_{\mathbf{z}} \log p_\theta(\mathbf{o}|\mathbf{z}) \notag \\
    &= \nabla_{\mathbf{z}} \log p(\mathbf{z}) + \nabla_{\mathbf{z}} \log p_\theta(\mathbf{o}|\mathbf{s}).
\end{align}
Since the prior $p(\mathbf{z})$ is a pre-defined Gaussian distribution and the likelihood $p_\theta(\mathbf{o}|\mathbf{s})$ is also Gaussian, the score $\nabla_{\mathbf{z}} \log p_\theta(\mathbf{z}|\mathbf{o})$ can be easily computed. With the score of the noise posterior distribution, we can employ Langevin dynamics to sample from this posterior:
\begin{equation}
    \mathbf{z}_{k+1} = \mathbf{z}_k + \frac{\eta}{2} \nabla_{\mathbf{z}} \log p_\theta(\mathbf{z}_k|\mathbf{o}) + \sqrt{\eta} \cdot \mathbf{w}_k, \quad \mathbf{w}_k \sim \mathcal{N}(\mathbf{0}, \mathbf{I}),
\end{equation}
where k differentiates the iterations of the Langevin dynamics. To enhance the practical performance of this latent space Langevin sampling, we identified three key effective strategies through a series of comparative experiments.

\subsubsection{Increasing latent degrees of freedom}
We identified that the composition of the optimizable latent variables is critical for sampling fidelity. In our framework, the latent variable $\mathcal{Z}$ is defined as a composite high-dimensional set encompassing all sources of stochasticity: the initial input vectors for both the initial state and transition models, as well as the noise vectors injected into every layer of both networks. Crucially, to further enhance optimization flexibility, we expanded the channel dimension of these layer-wise noise injections from single-channel to multi-channel. This design significantly multiplies the optimizable parameter space, enabling fine-grained tuning at specific feature levels.

Optimization with low degrees of freedom (e.g., optimizing only initial noise) faces a dual challenge: a constrained search space that imposes a theoretical ceiling on solution quality, and a rugged energy landscape populated with local minima. In contrast, our joint optimization strategy mitigates the first challenge by expanding the search space and addresses the second by decomposing the highly coupled global optimization into multi-level cooperative tasks. This structural flexibility allows the optimizer to escape global stagnation by adjusting local noise variables, thereby accessing higher-quality solution regions inaccessible to global-only adjustments.

\subsubsection{Choice of activation function}
We further observed that the choice of non-linear activation functions also contributes to sampling fidelity. Empirical comparisons indicated that employing Tanh as the primary activation yields improved fitting accuracy compared to standard ReLU~\citep{nair2010rectified} or LeakyReLU~\citep{maas2013rectifier} configurations. Interestingly, we initially hypothesized that this benefit stemmed from avoiding the non-smooth ``kink'' of ReLUs at zero; however, preliminary exploratory experiments did not support this. Consequently, we revised our hypothesis: Tanh's benefit likely arises from its property of providing a maximal gradient at the origin. Unlike ReLU or LeakyReLU, where gradients near zero can be small or abrupt, Tanh appears to offer a more favorable driving force for our gradient-based latent optimization process. This finding aligns with recent advances in Physics-Informed Neural Networks (PINNs), where smooth activation functions (e.g., Tanh) are preferred for modeling continuous physical quantities~\citep{Raissi2019}. Given these empirical benefits, we uniformly adopted Tanh throughout our model architecture.

\subsubsection{Choice of noise prior distribution}
\begin{figure}[!ht] 
    \centering

    \begin{minipage}[t]{0.03\linewidth}
        \vspace{0pt} \textbf{a}
    \end{minipage}
    \hfill
    \begin{minipage}[t]{0.96\linewidth}
        \vspace{0pt}
        \includegraphics[width=0.21\linewidth]{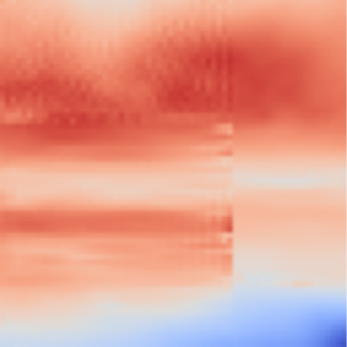}\hfill
        \includegraphics[width=0.21\linewidth]{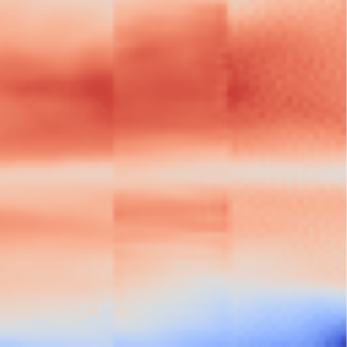}\hfill
        \includegraphics[width=0.21\linewidth]{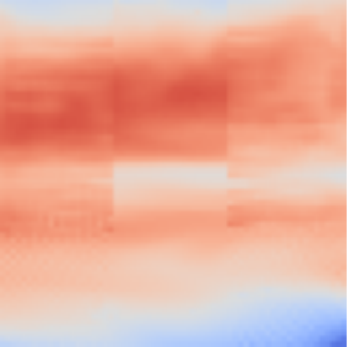}\hfill
        \includegraphics[width=0.21\linewidth]{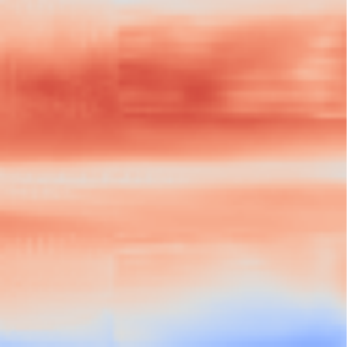}
    \end{minipage}

    \vspace{1em} 

    \begin{minipage}[t]{0.03\linewidth}
        \vspace{0pt} \textbf{b}
    \end{minipage}
    \hfill
    \begin{minipage}[t]{0.96\linewidth}
        \vspace{0pt}
        \includegraphics[width=0.21\linewidth]{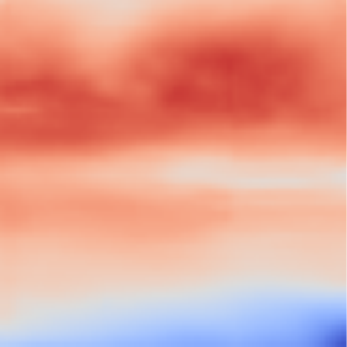}\hfill
        \includegraphics[width=0.21\linewidth]{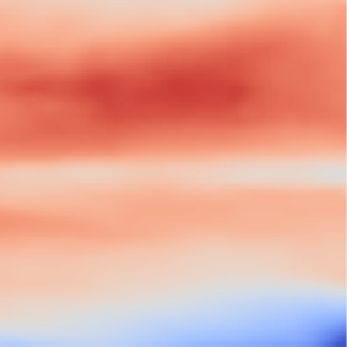}\hfill
        \includegraphics[width=0.21\linewidth]{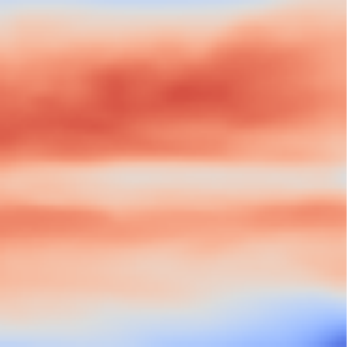}\hfill
        \includegraphics[width=0.21\linewidth]{./Gaussian-0.pdf}
    \end{minipage}

    \vspace{0.5em}

    \caption{
    Comparison of spatial continuity under different noise priors.
    a, Reconstructions using a uniform prior $\mathcal{U}(-1, 1)$ exhibit fracture-like artifacts.
    b, Reconstructions using a Gaussian prior $\mathcal{N}(0, \mathbf{I})$ demonstrate smooth structures.
    }
    \label{fig:noise_prior_comparison}
    \vspace{-3em}
\end{figure}

We further investigated the impact of the prior distribution type for the injected noise vectors $\boldsymbol{\varepsilon}$. We compared two standard choices: a Uniform distribution $\mathcal{U}(-1, 1)$ and a Standard Gaussian distribution $\mathcal{N}(0, \mathbf{I})$.
As visualized in Fig.~\ref{fig:noise_prior_comparison}, a striking contrast was observed. Employing a Uniform distribution (Fig.~\ref{fig:noise_prior_comparison}a) induces unnatural spatial discontinuities, particularly visible as vertical seams, thereby compromising the physical coherence of the reconstructed field. Conversely, the Gaussian noise prior (Fig.~\ref{fig:noise_prior_comparison}b) yields spatially smooth and physically consistent reconstructions, effectively eliminating these artifacts.

We attribute this phenomenon to the optimization dynamics in the latent space. The log-prior of a Uniform distribution has zero gradient within its support, failing to provide effective regularization guidance for unobserved regions. In contrast, the Gaussian prior imposes a smooth quadratic penalty ($\propto -\|\boldsymbol{\varepsilon}\|^2$), providing continuous gradient feedback that regularizes the latent variables. This property is highly compatible with our gradient-based Langevin sampling framework and is consistent with the standard assumption of Gaussian background error covariance in optimal interpolation and variational data assimilation~\citep{Lorenc1986, Bennett2002}. Consequently, we utilize Gaussian noise for all injected noise layers.

\subsection{M-step: Model evolution via multi-objective learning}
\label{subsec:mstep}

In the M-step, we update the model parameters $\theta = \{\theta_{\text{init}}, \theta_{\text{trans}}\}$ by maximizing the expected log-likelihood of the complete data~\citep{Dempster1977EM}. We utilize the high-fidelity state trajectories $\hat{\mathbf{s}} = \{\hat{\mathbf{s}}_0, \dots, \hat{\mathbf{s}}_T\}$ reconstructed in the E-step as pseudo-ground-truth training data. Given the architectural orthogonality between the initial state distribution and the temporal dynamics, we decouple the global optimization into two parallel sub-tasks:

It is worth noting that we employ different adversarial loss formulations for the two sub-tasks. For the Initial State Model, we adopt the non-saturating loss with R1 and PL regularization as recommended by StyleGAN2-ADA~\citep{Karras2020ada}. This choice is motivated by its proven stability and effectiveness in training high-fidelity unconditional generative models. For the State Transition Model, we utilize the original min-max adversarial loss from the conditional GAN framework~\citep{goodfellow2014generative, mirza2014conditional}, consistent with the foundational Pix2Pix~\citep{isola2017image} approach for image-to-image translation tasks. This deliberate choice allows each component to leverage the most established and robust training strategy for its specific task.

\subsubsection{Initial state model update}
\label{sssec:init_update}

We optimize the parameters $\theta_{\text{init}}$ (encompassing both the generator $G_{\theta_{\text{init}}}$ and discriminator $D_{\theta_{\text{init}}}$) to match the distribution of the reconstructed initial states $\hat{\mathbf{s}}_0$.

The generator objective minimizes the adversarial loss combined with path length regularization:
\begin{equation}
    \mathcal{L}_{G} = \mathbb{E}_{\mathbf{z}} \left[ \text{softplus}\left( -D_{\theta_{\text{init}}}(G_{\theta_{\text{init}}}(\mathbf{z})) \right) \right] + \lambda_{pl} \mathcal{L}_{pl}.
\end{equation}
where $\mathcal{L}_{pl}$ denotes the Path Length (PL) regularization~\citep{karras2020analyzing}. This term encourages a smooth latent space, essential for the stability of the gradient-based Langevin sampling in the E-step.

The discriminator objective maximizes the distinction between real and generated images, stabilized by the R1 gradient penalty:
\begin{equation}
    \mathcal{L}_{D} = \mathbb{E}_{\hat{\mathbf{s}}_0} \left[ \text{softplus}\left( -D_{\theta_{\text{init}}}(\hat{\mathbf{s}}_0) \right) \right] + \mathbb{E}_{\mathbf{z}} \left[ \text{softplus}\left( D_{\theta_{\text{init}}}(G_{\theta_{\text{init}}}(\mathbf{z})) \right) \right] + \lambda_{R1} \mathcal{L}_{R1}.
\end{equation}
where $\mathcal{L}_{R1} =  \mathbb{E}_{\hat{\mathbf{s}}_0}[\|\nabla D_{\theta_{\text{init}}}(\hat{\mathbf{s}}_0)\|^2]$ penalizes the gradient norm on real data to prevent mode collapse.

\subsubsection{State Transition Model Update}
\label{sssec:trans_update}

The transition parameters $\theta_{\text{trans}}$ (parameterizing the Stochastic U-Net) are optimized to capture the stochastic physical dynamics across the entire temporal sequence. We construct the training set by extracting all valid transition pairs $\{(\hat{\mathbf{s}}_t, \hat{\mathbf{s}}_{t+1})\}_{t=0}^{T-1}$ from the high-fidelity trajectories reconstructed in the E-step. The optimization is formulated as a min-max game over the cumulative loss across all time steps:
\begin{equation}
    \min_{G_{\theta_{\text{trans}}}} \max_{D_{\theta_{\text{trans}}}} \mathcal{L}_{\text{trans}} = \sum_{t=0}^{T-1} \left( \mathcal{L}_{\text{cGAN}}^{(t)} + \lambda \mathcal{L}_{L1}^{(t)} \right).
    \label{eq:chaocan}
\end{equation}
where the conditional adversarial term $\mathcal{L}_{\text{cGAN}}^{(t)}$~\citep{mirza2014conditional} ensures that the predicted transition from $\hat{\mathbf{s}}_t$ to $\hat{\mathbf{s}}_{t+1}$ is distributionally indistinguishable from real physical evolution:
\begin{equation}
    \mathcal{L}_{\text{cGAN}}^{(t)} = \mathbb{E}_{(\hat{\mathbf{s}}_t, \hat{\mathbf{s}}_{t+1})} [\log D_{\theta_{\text{trans}}}(\hat{\mathbf{s}}_t, \hat{\mathbf{s}}_{t+1})] + \mathbb{E}_{\hat{\mathbf{s}}_t, \mathbf{z}} [\log(1 - D_{\theta_{\text{trans}}}(\hat{\mathbf{s}}_t, G_{\theta_{\text{trans}}}(\hat{\mathbf{s}}_t, \mathbf{z})))].
\end{equation}
To further enforce pixel-wise fidelity of the prediction, we incorporate the L1 reconstruction loss:
\begin{equation}
    \mathcal{L}_{L1}^{(t)} = \mathbb{E}_{(\hat{\mathbf{s}}_t, \hat{\mathbf{s}}_{t+1}), \mathbf{z}} [\| \hat{\mathbf{s}}_{t+1} - G_{\theta_{\text{trans}}}(\hat{\mathbf{s}}_t, \mathbf{z}) \|_1].
\end{equation}
Here, $\mathbf{z}$ denotes the injected noise tensor modeling prediction uncertainty. In Eq.~\eqref{eq:chaocan}, $\lambda$ balances perceptual realism with structural accuracy.

\section{Results}
\label{sec:results}

\subsection{A Generative State-Space Modeling Framework for Incomplete Observations}
\label{subsec:framework_overview}

\begin{figure}[htbp] 
    \centering
    \includegraphics[width=0.85\linewidth]{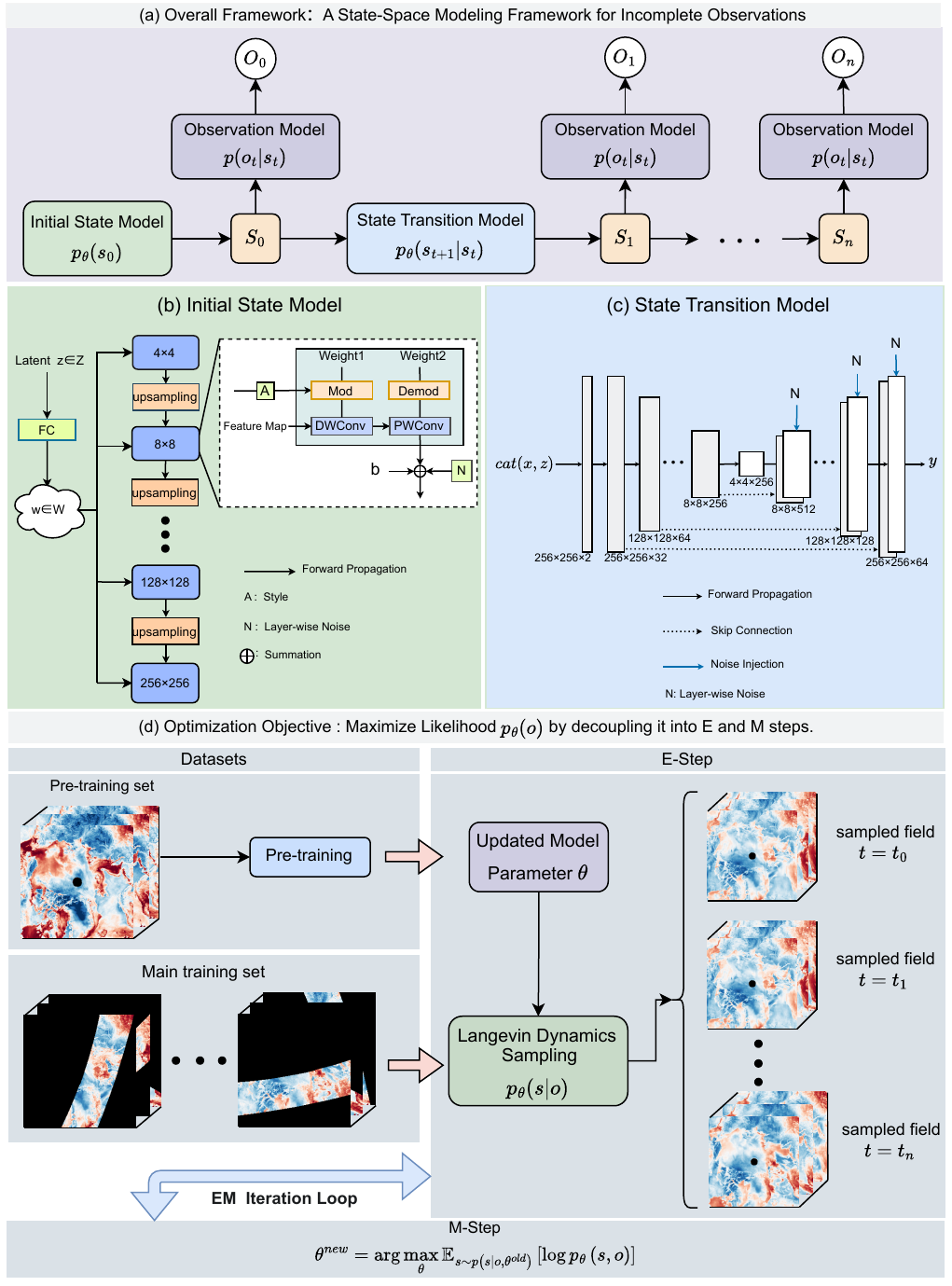} 
    \caption{\small Schematic of the generative state-space modeling framework.
    \textbf{a}, Probabilistic graphical model illustrating the hidden Markov process. Latent complete physical states ($S$) evolve temporally and emit partial observations ($O$) via the observation model. 
    \textbf{b}, \textbf{c}, Neural architectures for the generator components. 
    \textbf{b}, The Initial State Model ($p_\theta(\mathbf{s}_0)$) utilizes a StyleGAN2-ADA backbone with modulated depthwise separable convolutions (DWConv + PWConv) to map latent code $\mathbf{w}$ to spatial fields. 
    \textbf{c}, The State Transition Model ($p_\theta(\mathbf{s}_{t+1}|\mathbf{s}_t)$) employs a stochastic U-Net that processes concatenated inputs ($\text{cat}(\mathbf{x}, \mathbf{z})$) with multi-scale skip connections and noise injection ($N$). 
    \textbf{d}, Optimization workflow based on the Expectation-Maximization (EM) algorithm. The pipeline begins with model initialization using a pre-training dataset. In the main loop, raw satellite swath data serve as the primary training dataset. The E-step infers the posterior of hidden states ($S$) via Langevin dynamics sampling (Sampled Field), which are subsequently used in the M-step to update model parameters $\theta$ according to the maximization objective shown in the bottom panel.}
    \label{fig:framework}
\end{figure}

To address the challenge of capturing complex ocean physical dynamics from sparse satellite swath data, we constructed a data-driven generative state-space modeling framework (Fig.~\ref{fig:framework}a). Targeting the rigorous demands of high-resolution ocean simulation, we implemented tailored designs within the model architecture. First, the Initial State Model (Fig.~\ref{fig:framework}b) adopts a lightweight modified StyleGAN2-ADA~\citep{Karras2020ada}, incorporating a Depthwise Separable Convolution (DWConv) strategy~\citep{Chollet2017} to significantly reduce computational complexity while maintaining generation quality. Second, the State Transition Model (Fig.~\ref{fig:framework}c) utilizes a Stochastic U-Net~\citep{Ronneberger2015, Kohl2018ProbUNet}; by injecting random noise at the decoder levels, it explicitly captures the inherent uncertainties within the physical evolution process. Complementing these generative components, we defined an explicit probabilistic Observation Model to bridge the generated states with real-world data, formulating observations as sparse measurements with Gaussian noise.

Furthermore, to effectively handle the intrinsic cyclic dependency between "model learning" (which requires complete data) and "state reconstruction" (which requires a trained model), we formalized the training process as an iterative optimization framework based on the Expectation-Maximization (EM) algorithm~\citep{Brajard2020JCS} (Fig.~\ref{fig:framework}d). Unlike traditional incremental learning, this system achieves iterative improvement directly from spatiotemporally varying sparse observational data by alternating between the E-step and the M-step.

\subsection{Experimental design and dual-track validation strategy}
\label{subsec:exp_design}

\subsubsection{Experimental setup and datasets}
\label{sssec:setup_data}
To comprehensively evaluate performance across controlled and operational scenarios, we implemented a dual-track validation strategy. For rigorous quantitative assessment, we established an Observing System Simulation Experiment (OSSE) framework derived from the CMIP6 BCC-CSM2-MR climate model~\citep{Wu2019BCC}, treating daily sea surface temperature (tos) fields as the ground truth~\citep{Griffies2016GMD}. Synthetic incomplete observations were generated by applying structural masks---occluding approximately 67\% of the spatial domain---superimposed with Gaussian noise to mimic sensor imperfections. 

To assess operational generalization, we employed real-world Level-1 swath data from the FY-3D Microwave Radiation Imager (MWRI), focusing on the Arctic region ($>66^\circ$N). In contrast to the standardized simulation grid, this dataset is characterized by high-frequency yet non-uniform temporal sampling ($\sim$1.72\,h intervals) driven by polar orbital dynamics, presenting naturally occurring along-track data voids without ground-truth references.

\subsubsection{Progressive training protocol}
\label{sssec:training}
To bridge the gap between limited data availability and high-fidelity modeling requirements, we implemented a rigorous two-stage strategy analogous to curriculum learning. The process commences with a warm-up phase, where the generative models are initialized via supervised pre-training on a limited subset of complete physical fields (approx. 800 samples), thereby establishing a robust prior over spatial statistics. Subsequently, the framework transitions to the core EM-based iterative learning phase utilizing a larger corpus of exclusively incomplete observations (approx. 1,500 samples). This transition enables the model to progressively internalize stochastic physical dynamics while simultaneously enhancing reconstruction fidelity through the alternating Expectation and Maximization steps, effectively unlocking the capability to learn directly from sparse data.

\subsubsection{Task formulation}
\label{sssec:task}
While our proposed framework is theoretically applicable to sequential problems of arbitrary length $T$, in this study, we focus on the fundamental unit of temporal evolution: the $T=1$ scenario, comprising two consecutive frames ($t=0, 1$). This setting allows us to balance computational costs while capturing core thermodynamic dynamics. Consequently, the specific experimental objective across both simulated and real-world scenarios is to reconstruct the complete system state $\mathbf{s} = \{\mathbf{s}_0, \mathbf{s}_1\}$ from the sparse observation frames $\mathbf{o} = \{\mathbf{o}_0, \mathbf{o}_1\}$.

\subsection{High-fidelity reconstruction and forecasting in simulation}
\label{subsec:simulation_results}

\subsubsection{Evaluation metrics}
\label{sssec:metrics}
To comprehensively assess model performance, we employed Root Mean Square Error (RMSE) ~\citep{Wilks2011}, Mean Absolute Error (MAE) ~\citep{Wilks2011}, and Structural Similarity Index (SSIM)~\citep{Wang2004} as evaluation metrics, capturing both pixel-level accuracy and structural fidelity. Crucially, to ensure physical relevance, all quantitative errors were calculated in degrees Celsius ($^\circ$C) within the valid physical range of 15--35\,$^\circ$C.

\subsubsection{Baseline model for comparison}
\label{sssec:baseline_model}
To quantitatively demonstrate the efficacy of our iterative EM-based learning framework, we establish a baseline model, hereafter referred to as the Pre-trained Model. This model is derived from the initial warm-up phase of our training protocol. Specifically, it is trained via standard supervised learning on a limited set of complete ground-truth fields. The Final Model, in contrast, is the result of the full iterative training process, which refines the pre-trained parameters by learning directly from incomplete observations. This comparison is designed to isolate and quantify the performance gains attributable to the EM-driven learning from sparse data.

\subsubsection{High-fidelity reconstruction under structural masks}
\label{sssec:recon_structural}

Our central hypothesis is that the proposed Expectation--Maximization (EM) framework enables the model to bootstrap accurate physical representations from severely incomplete observations. Within an Observing System Simulation Experiment (OSSE), we take daily sea surface temperature (SST) fields from CMIP6 simulations as ground truth. Imperfect inputs are generated by applying structural masks together with additive Gaussian noise, emulating the irregular coverage characteristic of satellite retrievals.

Training follows a sequential split with two evaluated checkpoints. First, a \textbf{Pre-trained} baseline is obtained from a short warm-up stage on an early subset of complete, unmasked SST fields, which initializes the network and establishes basic spatiotemporal dynamics. The pipeline then proceeds to the main stage on the remaining data, where inputs are \emph{exclusively} masked and noise-corrupted to mimic observation-like incompleteness. Starting from this baseline, we run 17 EM iterations: in each E-step, latent variables are refined via Langevin dynamics to reconstruct full spatiotemporal states; in each M-step, model parameters are updated with Adam using these reconstructions as targets. The \textbf{Final model} denotes the parameters after completing these EM iterations.

For evaluation, both checkpoints are tested under five structural occlusion patterns that remove the top, bottom, center, left, or right two-thirds of the spatial domain; reported metrics are averages across these scenarios. Quantitative results in Table~\ref{tab:reconstruction_metrics} support the hypothesis: relative to the pre-trained baseline, the final model yields marked gains across deterministic scores for reconstructing the initial state $\mathbf{s}_0$ and the one-step-ahead forecast $\mathbf{s}_1$.

\begin{figure}[htbp] 
    \centering
    \sffamily\small 
    \setlength{\tabcolsep}{1pt} 
    
    \newcolumntype{M}{>{\centering\arraybackslash}m{0.13\linewidth}}
    \newcolumntype{L}{>{\centering\arraybackslash}m{0.03\linewidth}} 
    \newcolumntype{C}{>{\centering\arraybackslash}m{0.04\linewidth}} 

    \begin{tabular}{ L M M M M M M C }
        ~ & \textbf{Obs.} & \textbf{GT} & \textbf{Baseline} & \textbf{Base.~Err.} & \textbf{Final Model } & \textbf{Final Model ~Err.} & \textbf{} \\
        \multicolumn{8}{c}{\vspace{0.05cm}} \\ 

        \scriptsize\bfseries\rotatebox{90}{t0} & 
        \includegraphics[width=\linewidth]{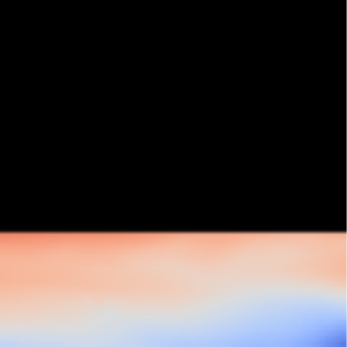} &
        \includegraphics[width=\linewidth]{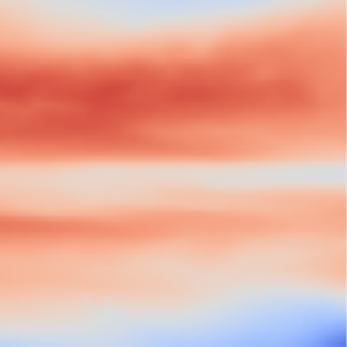} &
        \includegraphics[width=\linewidth]{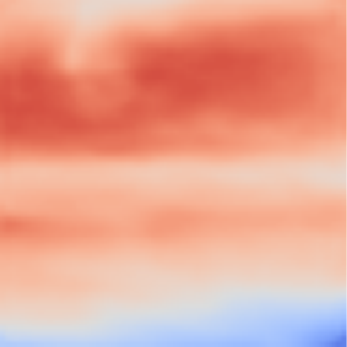} &
        \includegraphics[width=\linewidth]{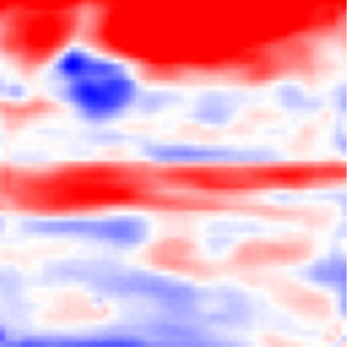} &
        \includegraphics[width=\linewidth]{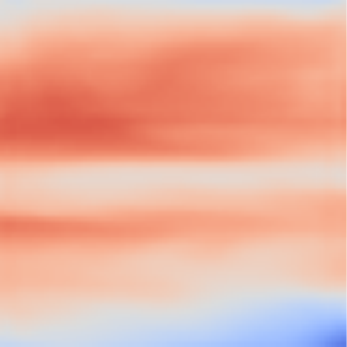} &
        \includegraphics[width=\linewidth]{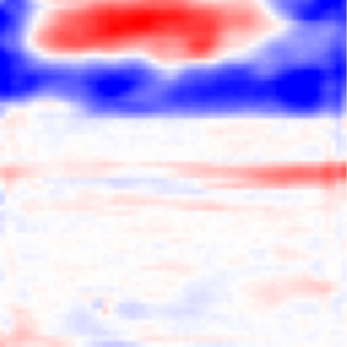} &
        \includegraphics[height=2.1cm, keepaspectratio]{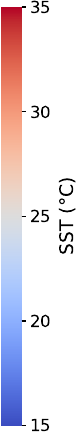} \\ 
        
        \scriptsize\bfseries\rotatebox{90}{t1} & 
        \includegraphics[width=\linewidth]{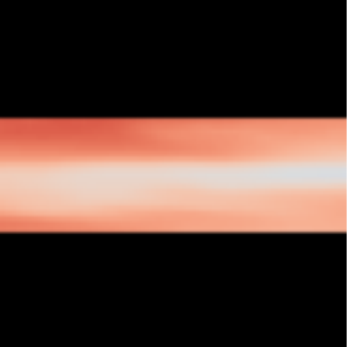} &
        \includegraphics[width=\linewidth]{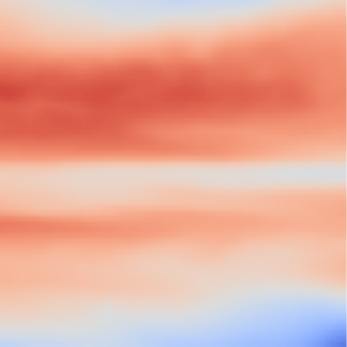} &
        \includegraphics[width=\linewidth]{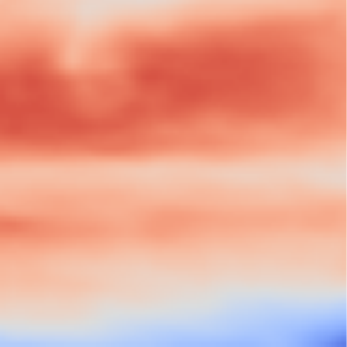} &
        \includegraphics[width=\linewidth]{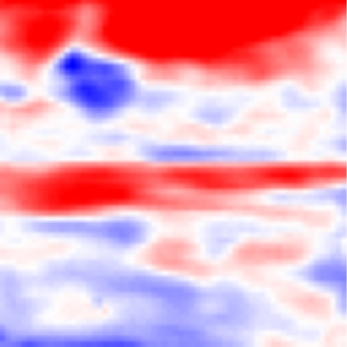} &
        \includegraphics[width=\linewidth]{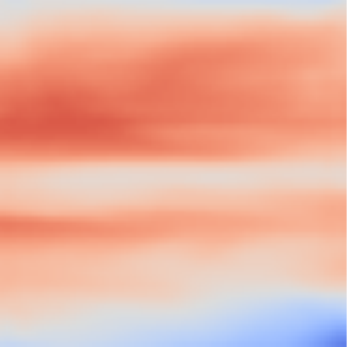} &
        \includegraphics[width=\linewidth]{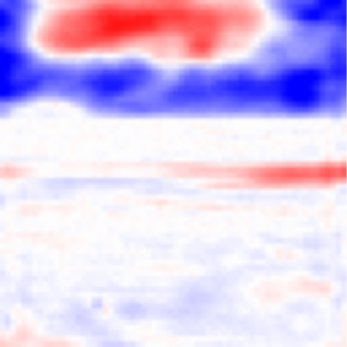} &
        \includegraphics[height=2.1cm, keepaspectratio]{./sim-verticle.pdf} \\

        \multicolumn{8}{c}{\vspace{0.2cm}} \\

        \scriptsize\bfseries\rotatebox{90}{t0} & 
        \includegraphics[width=\linewidth]{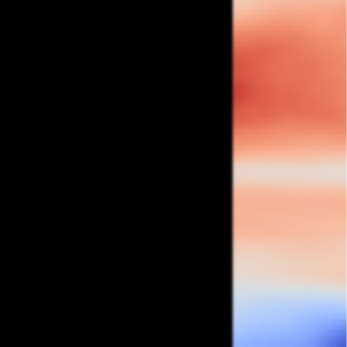} &
        \includegraphics[width=\linewidth]{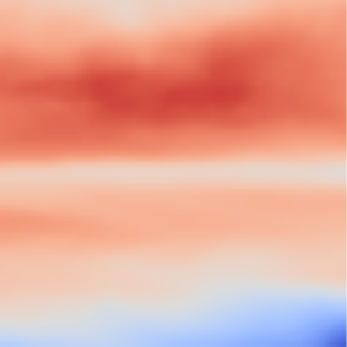} &
        \includegraphics[width=\linewidth]{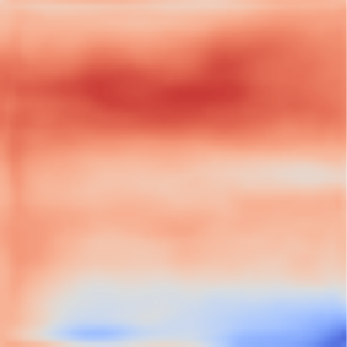} &
        \includegraphics[width=\linewidth]{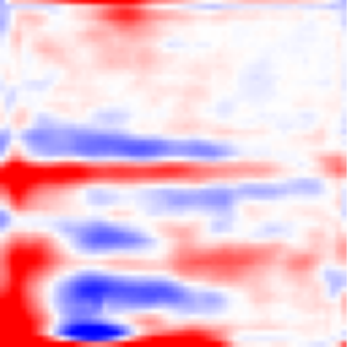} &
        \includegraphics[width=\linewidth]{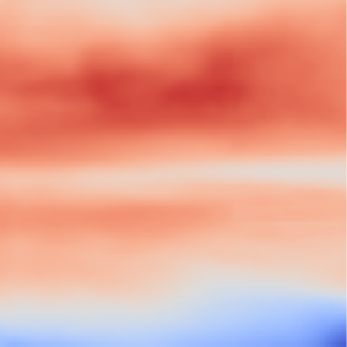} &
        \includegraphics[width=\linewidth]{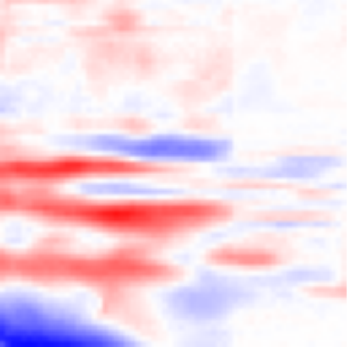} &
        \includegraphics[height=2.1cm, keepaspectratio]{./sim-verticle.pdf} \\ 
        
        \scriptsize\bfseries\rotatebox{90}{t1} & 
        \includegraphics[width=\linewidth]{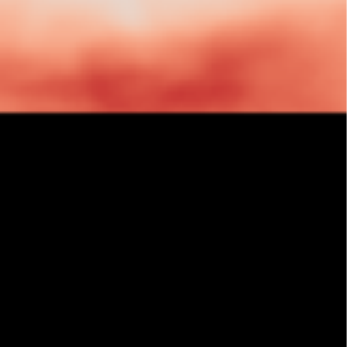} &
        \includegraphics[width=\linewidth]{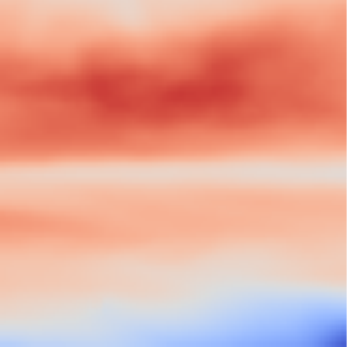} &
        \includegraphics[width=\linewidth]{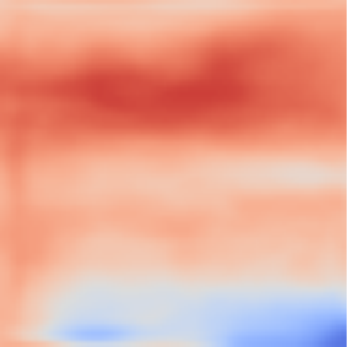} &
        \includegraphics[width=\linewidth]{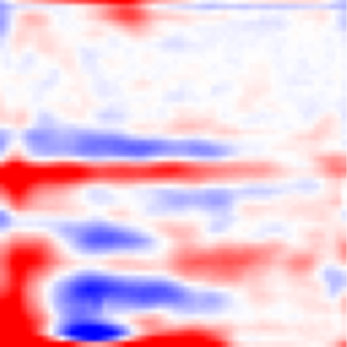} &
        \includegraphics[width=\linewidth]{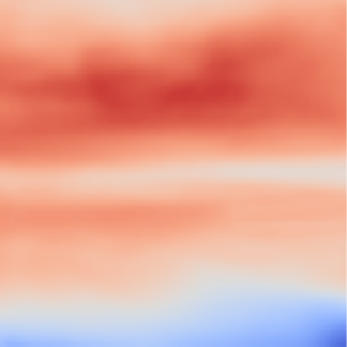} &
        \includegraphics[width=\linewidth]{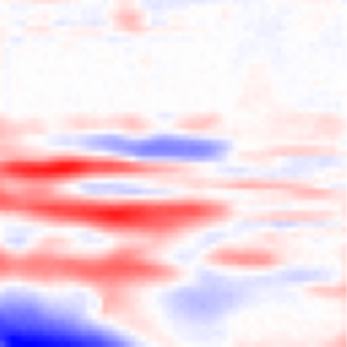} &
        \includegraphics[height=2.1cm, keepaspectratio]{./sim-verticle.pdf} \\

       \multicolumn{8}{c}{\vspace{0.2cm}} \\

        \scriptsize\bfseries\rotatebox{90}{t0} & 
        \includegraphics[width=\linewidth]{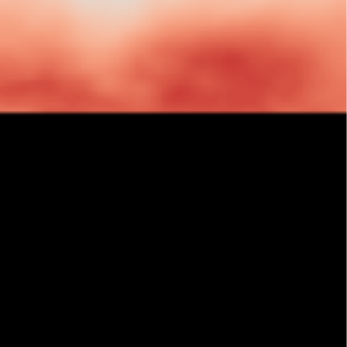} &
        \includegraphics[width=\linewidth]{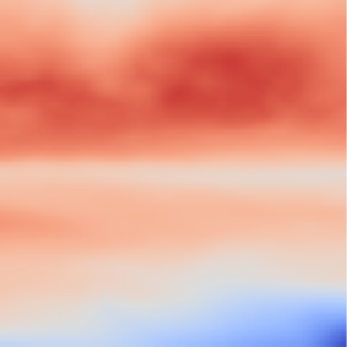} &
        \includegraphics[width=\linewidth]{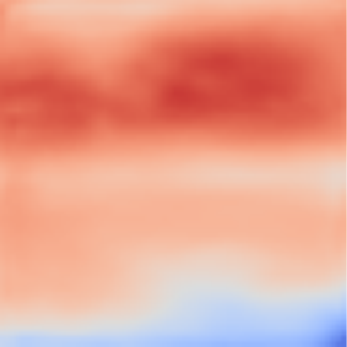} &
        \includegraphics[width=\linewidth]{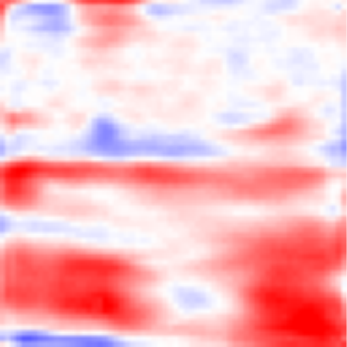} &
        \includegraphics[width=\linewidth]{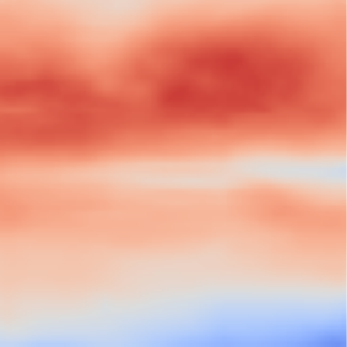} &
        \includegraphics[width=\linewidth]{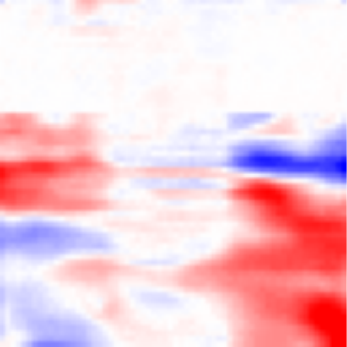} &
        \includegraphics[height=2.1cm, keepaspectratio]{./sim-verticle.pdf} \\ 
        
        \scriptsize\bfseries\rotatebox{90}{t1} & 
        \includegraphics[width=\linewidth]{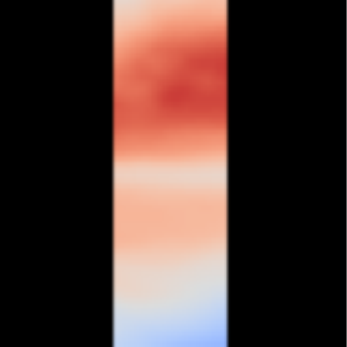} &
        \includegraphics[width=\linewidth]{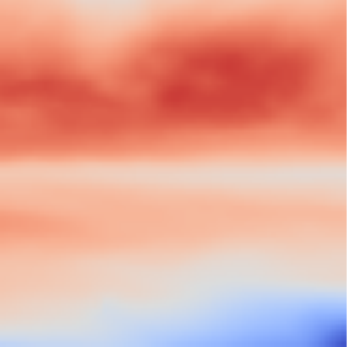} &
        \includegraphics[width=\linewidth]{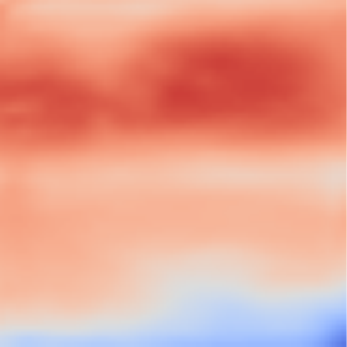} &
        \includegraphics[width=\linewidth]{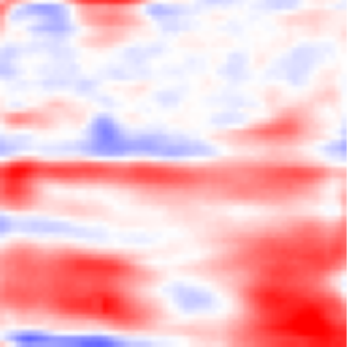} &
        \includegraphics[width=\linewidth]{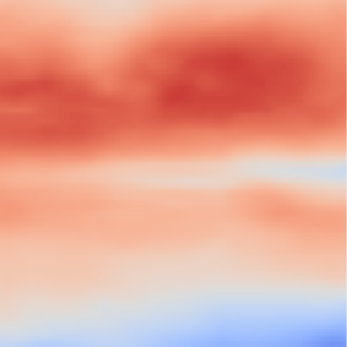} &
        \includegraphics[width=\linewidth]{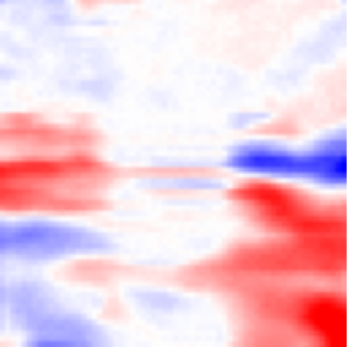} &
        \includegraphics[height=2.1cm, keepaspectratio]{./sim-verticle.pdf} \\
        
    \end{tabular}

    \caption{Qualitative comparison on representative simulation samples.
    Columns from left to right: Sparse Observation, Ground Truth, Baseline prediction, Baseline Error, Final Model  output, Final Model Error, and Scale. The reduction in residual intensity in the final column highlights the model's ability to recover fine-scale structures.}
    \label{fig:results_with_colorbar}
\end{figure}

\begin{table}[htbp] 
    \centering
    \caption{Quantitative evaluation on the simulated dataset. The table demonstrates the performance improvement of state reconstruction ($\mathbf{s}_0$) and forecasting ($\mathbf{s}_1$) via EM iteration. RMSE and MAE are reported in degrees Celsius ($^\circ$C).}
    \label{tab:reconstruction_metrics}
    \begin{tabular}{lcccc}
        \toprule
        \textbf{Model} & \textbf{States} & \textbf{RMSE ($^\circ$C) $\downarrow$} & \textbf{MAE ($^\circ$C) $\downarrow$} & \textbf{SSIM $\uparrow$} \\
        \midrule
        \multirow{2}{*}{Pre-trained Model} 
        & $\mathbf{s}_0$ & 0.9696 & 0.7034 & 0.9014 \\
        & $\mathbf{s}_1$ & 0.7587 & 0.4804 & 0.9138 \\
        \midrule
        \multirow{2}{*}{\textbf{Final Model}} 
        & $\mathbf{s}_0$ & \textbf{0.6969} & \textbf{0.4621} & \textbf{0.9355} \\
        & $\mathbf{s}_1$ & \textbf{0.5532} & \textbf{0.3197} & \textbf{0.9496} \\
        \bottomrule
    \end{tabular}
\end{table}

Specifically, for the reconstruction of the initial state $\mathbf{s}_0$, the Final Model reduced the RMSE by 28.13\% and the MAE by 34.30\%, alongside a significant improvement in SSIM. This indicates that the EM iteration enables the model to learn a more realistic prior distribution of physical fields, thereby enhancing its data assimilation capability. The performance improvement for $\mathbf{s}_1$ is equally significant, with an RMSE reduction of 27.09\% and an MAE reduction of 33.45\%. Since the reconstruction of $\mathbf{s}_1$ relies heavily on the physical evolution from $\mathbf{s}_0$ (governed by the State Transition Model), this result strongly evidences that the EM framework significantly improves the model's precision in simulating physical dynamics.

An interesting phenomenon observed is that the reconstruction errors for $\mathbf{s}_1$ are consistently lower than those for $\mathbf{s}_0$. We hypothesize that this is because the reconstruction of $\mathbf{s}_1$ is constrained by the physical evolution from $\mathbf{s}_0$, whereas $\mathbf{s}_0$ lacks constraints from past physics.

To complement quantitative metrics, we examined representative reconstruction results under structural masking conditions (\textbf{Fig. \ref{fig:results_with_colorbar}}). The Pre-trained Model, while capturing global spatial patterns, exhibits a tendency towards over-smoothing in unobserved regions. This results in a loss of high-frequency physical details, as reflected by higher residual magnitudes in the error maps. Conversely, the Final Model recovers sharper and physically more coherent structures, including distinct sub-mesoscale eddies~\citep{McWilliams2016} and temperature fronts. The visible reduction in error magnitude across the masked regions suggests that the EM-driven iterative process effectively corrects the inference bias present in the initialization phase, leading to representations that are closer to the ground truth~\citep{Ross2023JAMES}.

\subsubsection{Forecasting capability of the State Transition Model}
\label{sssec:forecasting_results}


Beyond assessing the coupled reconstruction performance, we further isolated and evaluated the pure forecasting capability of the State Transition Model ($G_{\theta_{\text{trans}}}$) itself. To this end, we designed a standard one-step forecasting experiment.

In this setup, we directly used the complete, noise-free ground-truth states $\mathbf{s}_0$ from the simulated dataset as input.
We then fed these perfect initial conditions into the state transition model to generate a one-step prediction, $\hat{\mathbf{s}}_1 = G_{\theta_{\text{trans}}}(\mathbf{s}_0)$.
Comparing $\hat{\mathbf{s}}_1$ with the true subsequent state $\mathbf{s}_1$ isolates the transition model's skill in simulating physical dynamics.
Table~\ref{tab:forecasting_metrics} summarizes one-step forecasting performance when $\mathbf{s}_0$ is taken from the ground truth.

\begin{table}[htbp]
  \centering
  \caption{Performance evaluation of the one-step forecasting task. The input is the ground-truth state $\mathbf{s}_0$. Best results are in bold.}
  \label{tab:forecasting_metrics}
  \begin{tabular}{lccc}
    \toprule
    \textbf{Model} & \textbf{RMSE ($^\circ$C) $\downarrow$} & \textbf{MAE ($^\circ$C) $\downarrow$} & \textbf{SSIM $\uparrow$} \\
    \midrule
    Pre-trained model & 0.9263 & 0.7288 & 0.9306 \\
    \textbf{Final model}    & \textbf{0.7431} & \textbf{0.5479} & \textbf{0.9408} \\
    \bottomrule
  \end{tabular}
\end{table}

The results in Table~\ref{tab:forecasting_metrics} show that, even when provided with identical perfect initial conditions, the final model still outperforms the pre-trained model in forecasting accuracy.
Specifically, the final model achieves noticeable reductions in both RMSE and MAE, together with a modest improvement in SSIM.

This substantial performance gain provides strong evidence that our EM framework does more than just end-to-end data fitting; it genuinely enables the State Transition Model itself to distill more accurate and authentic physical evolution laws from the iterative process of reconstruction and refinement. This result confirms that the model, trained via our proposed method, evolves into a more powerful and physically consistent standalone predictor.

\subsection{Generalization to real-world satellite observations}
\label{subsec:real_world_generalization}

\begin{figure}[htbp] 
    \centering
    \sffamily 
    \footnotesize 

    \begin{minipage}{0.05\textwidth}~ \end{minipage}%
    \hfill
    \begin{minipage}[b]{0.20\textwidth} \centering \textbf{Observation} \end{minipage}%
    \hfill
    \begin{minipage}[b]{0.20\textwidth} \centering \textbf{Baseline} \end{minipage}%
    \hfill
    \begin{minipage}[b]{0.20\textwidth} \centering \textbf{Final Model } \end{minipage}%
    \hfill
    \begin{minipage}[b]{0.08\textwidth} ~ \end{minipage}
    
    \vspace{2pt}

    \begin{minipage}[c]{0.05\textwidth} \centering \textbf{T\textsubscript{0}} \end{minipage}%
    \hfill
    \begin{minipage}[c]{0.20\textwidth} \includegraphics[width=\linewidth]{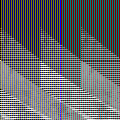} \end{minipage}%
    \hfill
    \begin{minipage}[c]{0.20\textwidth} \includegraphics[width=\linewidth]{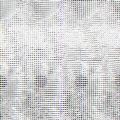} \end{minipage}%
    \hfill
    \begin{minipage}[c]{0.20\textwidth} \includegraphics[width=\linewidth]{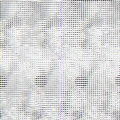} \end{minipage}%
    \hfill
    \begin{minipage}[c]{0.08\textwidth} \raisebox{-0.5\height}{\includegraphics[height=3.1cm]{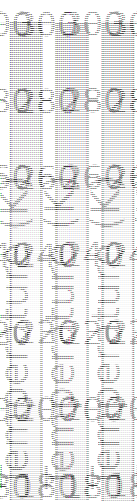}} \end{minipage}

    \vspace{1pt} 

    \begin{minipage}[c]{0.05\textwidth} \centering \textbf{T\textsubscript{1}} \end{minipage}%
    \hfill
    \begin{minipage}[c]{0.20\textwidth} \includegraphics[width=\linewidth]{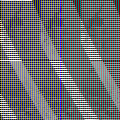} \end{minipage}%
    \hfill
    \begin{minipage}[c]{0.20\textwidth} \includegraphics[width=\linewidth]{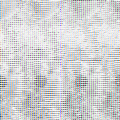} \end{minipage}%
    \hfill
    \begin{minipage}[c]{0.20\textwidth} \includegraphics[width=\linewidth]{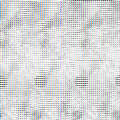} \end{minipage}%
    \hfill
    \begin{minipage}[c]{0.08\textwidth} \raisebox{-0.5\height}{\includegraphics[height=3.1cm]{./real-verticle.pdf}} \end{minipage}

    \vspace{2pt} 

    \begin{minipage}[c]{0.05\textwidth} \centering \textbf{T\textsubscript{0}} \end{minipage}%
    \hfill
    \begin{minipage}[c]{0.20\textwidth} \includegraphics[width=\linewidth]{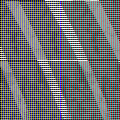} \end{minipage}%
    \hfill
    \begin{minipage}[c]{0.20\textwidth} \includegraphics[width=\linewidth]{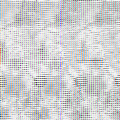} \end{minipage}%
    \hfill
    \begin{minipage}[c]{0.20\textwidth} \includegraphics[width=\linewidth]{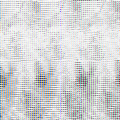} \end{minipage}%
    \hfill
    \begin{minipage}[c]{0.08\textwidth} \raisebox{-0.5\height}{\includegraphics[height=3.1cm]{real-verticle.pdf}} \end{minipage}
    
    \vspace{1pt}

    \begin{minipage}[c]{0.05\textwidth} \centering \textbf{T\textsubscript{1}} \end{minipage}%
    \hfill
    \begin{minipage}[c]{0.20\textwidth} \includegraphics[width=\linewidth]{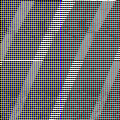} \end{minipage}%
    \hfill
    \begin{minipage}[c]{0.20\textwidth} \includegraphics[width=\linewidth]{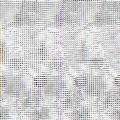} \end{minipage}%
    \hfill
    \begin{minipage}[c]{0.20\textwidth} \includegraphics[width=\linewidth]{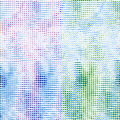} \end{minipage}%
    \hfill
    \begin{minipage}[c]{0.08\textwidth} \raisebox{-0.5\height}{\includegraphics[height=3.1cm]{real-verticle.pdf}} \end{minipage}

    \vspace{2pt} 

    \begin{minipage}[c]{0.05\textwidth} \centering \textbf{T\textsubscript{0}} \end{minipage}%
    \hfill
    \begin{minipage}[c]{0.20\textwidth} \includegraphics[width=\linewidth]{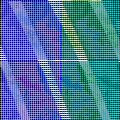} \end{minipage}%
    \hfill
    \begin{minipage}[c]{0.20\textwidth} \includegraphics[width=\linewidth]{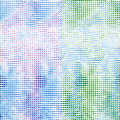} \end{minipage}%
    \hfill
    \begin{minipage}[c]{0.20\textwidth} \includegraphics[width=\linewidth]{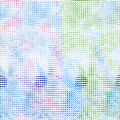} \end{minipage}%
    \hfill
    \begin{minipage}[c]{0.08\textwidth} \raisebox{-0.5\height}{\includegraphics[height=3.1cm]{real-verticle.pdf}} \end{minipage}
    
    \vspace{1pt}

    \begin{minipage}[c]{0.05\textwidth} \centering \textbf{T\textsubscript{1}} \end{minipage}%
    \hfill
    \begin{minipage}[c]{0.20\textwidth} \includegraphics[width=\linewidth]{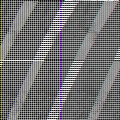} \end{minipage}%
    \hfill
    \begin{minipage}[c]{0.20\textwidth} \includegraphics[width=\linewidth]{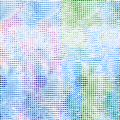} \end{minipage}%
    \hfill
    \begin{minipage}[c]{0.20\textwidth} \includegraphics[width=\linewidth]{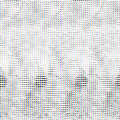} \end{minipage}%
    \hfill
    \begin{minipage}[c]{0.08\textwidth} \raisebox{-0.5\height}{\includegraphics[height=3.1cm]{real-verticle.pdf}} \end{minipage}

    \vspace{4pt} 

    \rmfamily 
    \caption{\small Robust spatiotemporal reconstruction from sparse satellite swaths. The figure compares raw observations, pre-trained baseline, and final model results across consecutive time steps for three samples. The framework ensures temporal continuity and recovers fine-scale structures.}
    \label{fig:real_world_results}
\end{figure}

\subsubsection{Pre-training set construction for real-world data}
\label{sssec:real_data_setup}
Beyond the quantitative benchmarks on simulated datasets, validating robustness against intrinsic data voids is crucial for operational viability. We applied the framework to real-world Level-1 swath data from the FY-3D satellite, where data incompleteness stems from orbital scanning geometries rather than artificial masks. In the absence of complete ground truth, we prioritized a qualitative assessment. To establish the Pre-trained Model (Baseline) for this dataset, we constructed a pre-training set via a multi-step pipeline: raw satellite swaths were first georeferenced and projected onto a regular Arctic grid (north of 66$^\circ$N); then, the approximately 14 daily orbital passes were sequentially aggregated to form a complete daily composite; finally, to recover the high-frequency temporal dynamics consistent with the satellite's orbital cycle ($\sim$1.72\,h), these daily composites were temporally interpolated to generate a sequence of approximately 14 spatially complete fields per day~\citep{Griffies2016GMD}.

\subsubsection{Structural fidelity and texture recovery}
\label{sssec:structural_fidelity}
Fig.~\ref{fig:real_world_results} consistently reveals a marked enhancement in structural fidelity across all three distinct observation scenarios. The Pre-trained Model (middle column) exhibits a characteristic ``over-smoothing'' bias, producing fields that are spatially continuous but lack high-frequency definition. This is particularly evident in the blurred ice edges and the homogenized texture within the main ice pack. In contrast, the Final Model (right column) successfully recovers sharp thermal gradients and intricate structural patterns, such as fine ice filaments and fragmented floes, within the unobserved regions. This suggests that the model has not merely learned to replicate smoothed, interpolated states, but has learned to sample from the complex manifold of valid physical states.

\subsubsection{Spatiotemporal coherence and physical plausibility}
\label{sssec:spatiotemporal_coherence}
Furthermore, a crucial test lies in the model's ability to maintain dynamic consistency over time. A detailed inspection of the temporal evolution from $T_0$ to $T_1$ highlights the superior performance of the Final Model. While the Baseline's evolution of fine-scale features appears less coherent across the transition in the middle column, the Final Model renders a more physically plausible evolution across all scenarios. In the first case (right column), the ice formations exhibit a clear, coherent rotational motion from $T_0$ to $T_1$, accompanied by a slight downward drift. This high-fidelity dynamic reconstruction is not an isolated success, as similar physically consistent evolutions are observed in the other scenarios in the right column as well, aligned with known Arctic sea-ice rheology.

This robust spatiotemporal consistency across diverse observation geometries strongly evidences that through the cyclic interplay of E-step and M-step, the framework has successfully internalized the governing physical laws. The model has learned not merely to ``inpaint'' static images, but to simulate a dynamic trajectory, effectively avoiding the temporal flickering often associated with simple interpolation methods~\citep{Brajard2020JCS}. The ability to generalize these learned laws to novel, real-world scenarios highlights its immense potential for data-sparse operational oceanography~\citep{Bell2015OM}.

\section{Discussion and Conclusion}
\label{sec:discussion}

\subsection{Breaking the data dependency barrier}
The central achievement of this study is the establishment and validation of a generative state-space modeling framework capable of iterative learning directly from sparse, incomplete observations. In the rapidly evolving landscape of AI for Earth System Science (AI4ESS), the reliance on pristine, complete reanalysis datasets (e.g., ERA5) has long been a bottleneck, limiting models to the role of ``emulators'' rather than independent ``simulators.'' Our work fundamentally challenges this status quo. By demonstrating that an AI model can be trained end-to-end using only partial observations, we provide a tangible pathway toward autonomous AI Earth system modeling, liberating the field from the computational and resolution constraints imposed by traditional data assimilation systems.

\subsection{The efficacy of the EM paradigm}
Our experimental results provide compelling evidence that the Expectation-Maximization (EM) algorithm serves as a robust paradigm for addressing the inherent circular dependency between ``state reconstruction'' and ``model learning.'' As evidenced by the significant performance leap from the ``Pre-trained Model'' to the ``Final Model'' (Table~\ref{tab:reconstruction_metrics}), the framework successfully instigates a ``bootstrapping'' learning process. This is a critical finding: it implies that deep generative models possess the capacity to refine their understanding of physical dynamics by leveraging their own imperfect reconstructions. Through the iterative interplay of the E-step (probabilistic state inference) and the M-step (parameter optimization), the model converges from a coarse approximation to a high-performance state that captures intricate physical dynamics, effectively distilling coherent physical laws from fragmented data.

\subsection{Implications for future Earth system modeling}
This core discovery paves a new, viable path for the development of next-generation AI Earth system models. It demonstrates that perfect supervision is not a prerequisite for learning complex physics. Instead, by designing sophisticated iterative learning frameworks, we can extract knowledge directly from the noisy, imperfect observations of the real world. This capability is particularly transformative for regions or variables where high-quality reanalysis data is scarce or nonexistent (e.g., deep ocean variables or polar regions). By enabling models to learn from raw satellite or in-situ observations, we move closer to constructing independent, powerful physical process simulators that can complement, or potentially supersede, traditional numerical methods in specific domains.

\subsection{Limitations and future outlook}
While our framework shows immense promise, we dge certain limitations. First, the computational cost of the Langevin dynamics sampling in the E-step is non-trivial, potentially limiting scalability for high-frequency, global-scale operational forecasting. Future work will explore accelerating this process via variational inference~\citep{kingma2013auto} or distilled diffusion solvers~\citep{salimans2022progressive}. Second, while we achieve high fidelity, ensuring strict adherence to conservation laws (e.g., mass and energy conservation) remains a challenge for purely data-driven generative models. Integrating physical constraints (Physics-Informed Machine Learning) into our probabilistic framework represents a promising direction~\citep{Raissi2019}. Despite these challenges, this work stands as a foundational step towards a more data-efficient and physically grounded era of AI meteorology and oceanography.


\appendix 

\section{Derivation of the Generative State-Space Model}
\label{app:derivations} 

\subsection{Maximum Likelihood Estimation Objective}
Consider a dataset of partial observations $\mathbf{o} = (\mathbf{o}^{(1)}, \mathbf{o}^{(2)}, \dots, \mathbf{o}^{(N)})$. Here, the superscript within the parentheses denotes different samples in the dataset, rather than time points. It should be noted that each sample in the dataset may contain observations from multiple different time points. We treat different samples in the dataset as independent and identically distributed. Afterwards, the goal of Maximum Likelihood Estimation (MLE) is to find the optimal parameters $\theta^*$ that maximize the probability of the data:
\begin{equation}
	\theta^* = \arg\max_\theta p_\theta(\mathbf{o}) = \arg\max_\theta \prod_{i=1}^N p_\theta(\mathbf{o}^{(i)}).
\end{equation}
For simplicity, we maximize the log-likelihood to transform the product into a sum:
\begin{equation}
	\theta^* = \arg\max_\theta \log p_\theta(\mathbf{o}) = \arg\max_\theta \sum_{i=1}^N \log p_\theta(\mathbf{o}^{(i)}).
	\label{eq:loglikehood}
\end{equation}

In our framework, we model the joint probability of the system state and the observation. Therefore, the probability of the observation is actually a marginal probability. Hence, to get $p_\theta(o^{(i)})$ in Eq.~\eqref{eq:loglikehood}, we need to integrate over the state space:
\begin{equation}
    p_\theta(\mathbf{o}^{(i)}) = \int p_\theta(\mathbf{s}^{(i)},\mathbf{o}^{(i)}) \, \mathrm{d}\mathbf{s}^{(i)} = \int p_\theta(\mathbf{s}^{(i)}) p_\theta(\mathbf{o}^{(i)} \mid \mathbf{s}^{(i)}) \, \mathrm{d}\mathbf{s}^{(i)}.
    \label{eq:integral} 
\end{equation}
In the framework, $p_\theta(\mathbf{s}^{(i)})$ is modeled through sophisticated neural networks (i.e. initial state model and state transition model), making the integral difficult to calculate. Consequently, the objective defined in Eq.~\eqref{eq:loglikehood} can not be directly optimized. 

\subsection{Decomposition of the Log-Likelihood}
For the sake of discussion, let's single out and analyze a specific term $\log p_\theta(\mathbf{o}^{(i)})$ from Eq.~\eqref{eq:loglikehood}. In fact, $\log p_\theta(\mathbf{o}^{(i)})$ can be decomposed as follows:
\begin{equation}
	\log p_\theta(\mathbf{o}^{(i)}) = \log p_\theta(\mathbf{s}^{(i)}, \mathbf{o}^{(i)}) - \log p_\theta(\mathbf{s}^{(i)} \mid \mathbf{o}^{(i)}).
    \label{eq:A4}
\end{equation}
Furthermore, we introduce an auxiliary distribution $q(\mathbf{s}^{(i)} \mid \mathbf{o}^{(i)})$ satisfying:
\begin{equation}
	q(\mathbf{s}^{(i)} \mid \mathbf{o}^{(i)}) \ge 0, \quad \int q(\mathbf{s}^{(i)} \mid \mathbf{o}^{(i)}) \, \mathrm{d}\mathbf{s}^{(i)} = 1.
	\label{eq:auxiliary}
\end{equation}
Taking the expectation of the log likelihood in Eq.~\eqref{eq:A4} with respect to $q$, we obtain:
\begin{equation}
	\mathbb{E}_{q} [\log p_\theta(\mathbf{o}^{(i)})] = \mathbb{E}_q[\log p_\theta(\mathbf{s}^{(i)}, \mathbf{o}^{(i)})] - \mathbb{E}_q[\log p_\theta(\mathbf{s}^{(i)} \mid \mathbf{o}^{(i)})].
    \label{eq:expectation_identity}
\end{equation}
The term on the left-hand side of Eq.~\eqref{eq:expectation_identity} can be expanded as follows:
\begin{align}
    \mathbb{E}_{q} [\log p_\theta(\mathbf{o}^{(i)})] 
    &= \int q(\mathbf{s}^{(i)} \mid \mathbf{o}^{(i)}) \log p_\theta(\mathbf{o}^{(i)}) \, \mathrm{d}\mathbf{s}^{(i)} \notag \\
    &= \log p_\theta(\mathbf{o}^{(i)}) \int q(\mathbf{s}^{(i)} \mid \mathbf{o}^{(i)}) \, \mathrm{d}\mathbf{s}^{(i)} \notag \\
    &= \log p_\theta(\mathbf{o}^{(i)}).
    \label{eq:lhs_proof}
\end{align}
Here, we utilized the definition of $q$ in Eq.~\eqref{eq:auxiliary}. Combining the results from Eq.~\eqref{eq:expectation_identity} and Eq.~\eqref{eq:lhs_proof}, we obtain the decomposition of the marginal log-likelihood:
\begin{equation}
    \log p_\theta(\mathbf{o}^{(i)}) = \mathbb{E}_{q}\left[\log p_\theta(\mathbf{s}^{(i)}, \mathbf{o}^{(i)})\right] - \mathbb{E}_{q}\left[\log p_\theta(\mathbf{s}^{(i)} \mid \mathbf{o}^{(i)})\right].
    \label{eq:decomposition}
\end{equation}
The second term on the right-hand side of Eq.~\eqref{eq:decomposition} relates to the cross-entropy between the auxiliary distribution $q(\mathbf{s}^{(i)} \mid \mathbf{o}^{(i)})$ and the posterior $p_\theta(\mathbf{s}^{(i)} \mid \mathbf{o}^{(i)})$.
\subsection{Iterative Optimization via the EM Algorithm}
We employ an iterative optimization strategy. Let $\theta^{(n)}$ denote the model parameters at the $n$-th iteration. Since $q$ in Eq.~\eqref{eq:decomposition} can be any distribution, we can readily set it to be equal to the exact posterior under the current parameters:
\begin{equation}
    q(\mathbf{s}^{(i)} \mid \mathbf{o}^{(i)}) = p_{\theta^{(n)}}(\mathbf{s}^{(i)} \mid \mathbf{o}^{(i)}).
\end{equation}
In this situation, the second term on the right-hand side of Eq.~\eqref{eq:decomposition} becomes the entropy of $q$. According to Gibbs' Inequality, any update to $\theta$ will result in an increase in the second term on the right-hand side of Eq.~\eqref{eq:decomposition}. Therefore, if we can increase the expected value of the complete-data log-likelihood (the first term on the right-hand side of Eq.~\eqref{eq:decomposition}) by updating $\theta$, we will also be able to increase the log-likelihood of the observed data (the left-hand side of Eq.~\eqref{eq:decomposition}). Incorporating Eq.~\eqref{eq:loglikehood}, the M-step at iteration $n+1$ solves
\begin{equation}
    \theta^{(n+1)} = \arg\max_\theta \sum_{i=1}^{N} \mathbb{E}_{\mathbf{s}^{(i)} \sim q(\mathbf{s}^{(i)} \mid \mathbf{o}^{(i)})} \left[ \log p_\theta(\mathbf{s}^{(i)}, \mathbf{o}^{(i)}) \right].
    \label{eq:m_step_theoretical}
\end{equation}

\paragraph{Monte Carlo Approximation.}
Since the analytic formula of the auxiliary distribution $q$ is unknown, exact computation of the expectations in Eq.~\eqref{eq:m_step_theoretical} is infeasible, necessitating the employment of Monte Carlo integration. Specifically, for an observation sample $\mathbf{o}^{(i)}$, we can approximate the expectation by drawing samples from the auxiliary distribution:
\begin{equation}
    \mathbb{E}_{\mathbf{s}^{(i)} \sim q(\mathbf{s}^{(i)} \mid \mathbf{o}^{(i)})} [\log p_\theta(\mathbf{s}^{(i)}, \mathbf{o}^{(i)})] \approx \frac{1}{M} \sum_{j=1}^{M} \log p_\theta(\mathbf{s}^{(i, j)}, \mathbf{o}^{(i)}).
    \label{eq:monte}
\end{equation}
In Eq.~\eqref{eq:monte}, $\mathbf{s}^{(i,j)}$ denotes a sample drawn from $q(\mathbf{s}^{(i)} \mid \mathbf{o}^{(i)})$, and $M$ signifies the number of samples for approximation. Theoretically, the larger the value of $M$, the more accurate the approximation of the expectation. However, in our experiments, we found that satisfactory results can be achieved even when $M=1$. Therefore, to balance computational efficiency with accuracy, we set \textbf{$M=1$} in the implementation. By substituting Eq.~\eqref{eq:monte} back into Eq.~\eqref{eq:m_step_theoretical}, we can get
\begin{equation}
	\theta^{(n+1)} = \arg\max_\theta \frac{1}{M}\sum_{i=1}^{N}\sum_{j=1}^{M} \log p_\theta(\mathbf{s}^{(i,j)}, \mathbf{o}^{(i)}).
	\label{eq:mcem_objective_appendix}
\end{equation}
Since $M$ is always equal to 1 in the implementation, we can omit it for simplicity. In this case, Eq.~\eqref{eq:mcem_objective_appendix} further simplifies to
\begin{equation}
	\theta^{(n+1)} = \arg\max_\theta \sum_{i=1}^{N} \log p_\theta(\mathbf{s}^{(i)}, \mathbf{o}^{(i)}).
	\label{eq:final_objective}
\end{equation}
Eq.~\eqref{eq:final_objective} matches Eq.~\eqref{eq:mcem_objective} in the main text; the optimization objective remains Eq.~\eqref{eq:loglikehood}.
Here, $\mathbf{s}^{(i)}$ represents the high-fidelity physical fields sampled during the E-step. The superscript ``$(i)$'' corresponds to different instances in the dataset, ensuring that each system state trajectory is strictly paired with its specific observation sequence.

\subsection{E-Step and M-Step Decoupling} 
\label{subsec:decoupling}

\subsubsection{E-Step: Efficient State Sampling}
Consider the states and observations over a sequence of time steps $t = 0, 1, \dots, T$:
\begin{equation}
\mathbf{s} = (\mathbf{s}_0, \mathbf{s}_1, \dots, \mathbf{s}_T), \quad \mathbf{o} = (\mathbf{o}_0, \mathbf{o}_1, \dots, \mathbf{o}_T).
\end{equation}
The posterior distribution of $\mathbf{s}$ given $\mathbf{o}$ is given by Bayes' rule:
\begin{equation}
p_\theta(\mathbf{s} \mid \mathbf{o}) \propto p_\theta(\mathbf{o} \mid \mathbf{s}) \, p_\theta(\mathbf{s}).
\label{eq:posterior_bayes}
\end{equation}
Assuming conditional independence of observations, the likelihood factorizes as the product over all time steps:
\begin{equation}
p_\theta(\mathbf{o} \mid \mathbf{s}) = \prod_{t=0}^{T} p_\theta(\mathbf{o}_t \mid \mathbf{s}_t).
\label{eq:likelihood_factorization}
\end{equation}
Combining Eqs.~\eqref{eq:posterior_bayes} and~\eqref{eq:likelihood_factorization}, the gradient of the log-posterior with respect to the state $\mathbf{s}$ is:
\begin{equation}
\nabla_{\mathbf{s}} \log p_\theta(\mathbf{s} \mid \mathbf{o})
= \sum_{t=0}^{T} \nabla_{\mathbf{s}} \log p_\theta(\mathbf{o}_t \mid \mathbf{s}_t) + \nabla_{\mathbf{s}} \log p_\theta(\mathbf{s}).
\label{eq:grad_log_posterior}
\end{equation}

Directly utilizing Eq.~\eqref{eq:grad_log_posterior} to perform Langevin dynamics sampling in the $\mathbf{s}$-space is intractable due to the high dimensionality of the physical state $\mathbf{s}$ and the complexity of its prior distribution, which renders the computation of the posterior score function prohibitively difficult. Instead, we leverage the generative mapping $\mathbf{s} = \mathcal{G}(\mathbf{z})$ (where $\mathbf{z} = (\mathbf{z}_0, \dots, \mathbf{z}_T)$ is a lower-dimensional latent variable) and perform Langevin dynamics sampling in the latent space $\mathbf{z}$ to approximate the posterior $p_\theta(\mathbf{z} \mid \mathbf{o})$. The gradient of the log-posterior with respect to $\mathbf{z}$ is decomposed as:
\begin{equation}
\nabla_{\mathbf{z}} \log p_\theta(\mathbf{z} \mid \mathbf{o}) = \nabla_{\mathbf{z}} \log p_\theta(\mathbf{o} \mid \mathbf{z}) + \nabla_{\mathbf{z}} \log p_\theta(\mathbf{z}).
\label{eq:score_z}
\end{equation}
Expanding the observation likelihood term (where $\mathbf{s}_t$ is a function of $\mathbf{z}$):
\begin{equation} 
\nabla_{\mathbf{z}} \log p_\theta(\mathbf{o} \mid \mathbf{z}) = \sum_{t=0}^{T} \nabla_{\mathbf{z}} \log p_\theta(\mathbf{o}_t \mid \mathbf{s}_t).
\end{equation}
Thus, the score function with respect to the latent variable $\mathbf{z}$ is given by:
\begin{equation}
\nabla_{\mathbf{z}} \log p_\theta(\mathbf{z} \mid \mathbf{o}) = \sum_{t=0}^{T} \nabla_{\mathbf{z}} \log p_\theta(\mathbf{o}_t \mid \mathbf{s}_t) + \nabla_{\mathbf{z}} \log p_\theta(\mathbf{z}).
\label{eq:estep_final}
\end{equation}

\subsubsection{M-Step: Decoupling the Parameter Updates}
The M-step in Eq.~\eqref{eq:final_objective} maximizes the complete-data log-likelihood over the $N$ sampled trajectories $\{(\mathbf{s}^{(i)},\mathbf{o}^{(i)})\}_{i=1}^N$. We now factorize the joint probability $p_\theta(\mathbf{s}, \mathbf{o})$ based on our generative state-space model structure. For a single trajectory $(\mathbf{s}^{(i)},\mathbf{o}^{(i)}) = \{(\mathbf{s}_0^{(i)},\mathbf{o}_0^{(i)}),(\mathbf{s}_1^{(i)},\mathbf{o}_1^{(i)}) \dots, (\mathbf{s}_T^{(i)},\mathbf{o}_T^{(i)})\}$, the log-likelihood decomposes as:
\begin{equation}
	\log p_\theta(\mathbf{s}^{(i)}, \mathbf{o}^{(i)}) = \log p_{\theta_{\text{init}}}(\mathbf{s}_0^{(i)}) + \sum_{t=1}^{T} \log p_{\theta_{\text{trans}}}(\mathbf{s}_t^{(i)} \mid \mathbf{s}_{t-1}^{(i)}) + \sum_{t=0}^{T} \log p(\mathbf{o}_t^{(i)} \mid \mathbf{s}_t^{(i)}).
	\label{eq:decomposition_appendix}
\end{equation}
Since the observation likelihood terms $\log p(\mathbf{o}_t^{(i)} \mid \mathbf{s}_t^{(i)})$ are explicitly modeled by the Gaussian distribution and possess no learnable parameters, they are constant with respect to $\theta$. Therefore, these terms can be omitted in the M-step update.

Substituting the decomposition in Eq.~\eqref{eq:decomposition_appendix} into Eq.~\eqref{eq:final_objective}, and noting that $\theta_{\text{init}}$ and $\theta_{\text{trans}}$ are disjoint parameter sets ($\theta_{\text{init}}$ and $\theta_{\text{trans}}$ are encapsulated within the initial state model and the state transition model, respectively), the M-step update decouples into two independent sub-problems:
\begin{enumerate}
	\item \textbf{Initial State Model Update:}
	\begin{equation}
		\theta_{\text{init}}^{\text{new}} = \arg\max_{\theta_{\text{init}}} \sum_{i=1}^{N} \log p_{\theta_{\text{init}}}(\mathbf{s}_0^{(i)})
	\end{equation}
	This is equivalent to minimizing the negative log-likelihood, corresponding to the loss $\mathcal{L}_{\text{initial}}$ defined in the main text.
	
	\item \textbf{State Transition Model Update:}
	\begin{equation}
		\theta_{\text{trans}}^{\text{new}} = \arg\max_{\theta_{\text{trans}}} \sum_{i=1}^{N} \sum_{t=1}^{T} \log p_{\theta_{\text{trans}}}(\mathbf{s}_t^{(i)} \mid \mathbf{s}_{t-1}^{(i)})
	\end{equation}
	This corresponds to maximizing the conditional log-likelihood of the transitions, which maps to the loss $\mathcal{L}_{\text{transition}}$.
\end{enumerate}

Thus, the M-step effectively minimizes the combined loss $\mathcal{L}_M = \mathcal{L}_{\text{initial}} + \mathcal{L}_{\text{transition}}$.

\section{Open Research}

\subsection*{Data Availability Statement}

CMIP6 model output was obtained from the Earth System Grid Federation (ESGF) CMIP6 interface hosted at the Lawrence Livermore National Laboratory node (\url{https://esgf-node.llnl.gov/projects/cmip6/}). This study uses daily sea surface temperature (\texttt{tos}; CMIP6 table \texttt{Oday}) from the Beijing Climate Center Climate System Model version 2 with medium resolution (\texttt{BCC-CSM2-MR}) \texttt{historical} experiment, variant label \texttt{r1i1p1f1}, on the native ocean grid (\texttt{gn}).

Satellite data are FengYun-3D (\texttt{FY-3D}) Microwave Radiation Imager (\texttt{MWRI}) Level-1 (\texttt{L1}) products distributed by the National Satellite Meteorological Center (NSMC), China Meteorological Administration. Data access, product metadata, and distribution policies are available through the NSMC data portal: \url{https://data.nsmc.org.cn/DataPortal/cn/data/dataset.html?dataTypeCode=L1&satelliteCode=FY3D&instrumentTypeCode=MWRI}. Users must comply with NSMC registration requirements, data-use terms, and official citation guidance for the specific product version and observation period used in this study.

\subsection*{Software and Code Availability Statement}

The code used to implement the methods of this study is publicly available in the GitHub repository \url{https://github.com/kyy-logs/Incomplete-Observations-Boost-Evo--lutionary-Performance-in-Ocean-Modeling}. The software is distributed under the license terms provided in the repository file \texttt{LICENSE.txt}. Installation, dependencies, and containerized execution are described in the repository \texttt{README.md} and \texttt{Dockerfile}.

\section*{Competing interests}
The authors declare that they have no known competing financial interests or personal relationships that could have appeared to influence the work reported in this paper.

\section*{Acknowledgements}
This work was supported by the Natural Science Foundation of China (Grant No. 42406192), the Fundamental Research Funds for the Central Universities (Grant No. 202413040), the National Science and Technology Major Project of China (Grant No. 2022ZD0117201), the Key R\&D Program of Shandong Province (Grant No. 2025CXPT185), and the Postdoctoral Project of Qingdao (Grant No. QDBSH20240102021). 

We dge the World Climate Research Programme (WCRP) and the Climate Model Intercomparison Project (CMIP6) for providing the climate model outputs. We thank the Beijing Climate Center (BCC) for making the BCC-CSM2-MR historical simulations available through the Earth System Grid Federation (ESGF, \url{https://esgf-node.llnl.gov/projects/cmip6/}). We also thank the National Satellite Meteorological Center (NSMC), China Meteorological Administration, for providing the FengYun-3D (FY-3D) Microwave Radiation Imager (MWRI) Level-1 data (\url{https://data.nsmc.org.cn}).

\bibliography{cas-refs} 

\end{document}